\documentclass[english]{article}
\usepackage{geometry}

\usepackage[T1]{fontenc}
\usepackage{iftex}
\ifPDFTeX
\usepackage[latin9]{inputenc}
\fi
\usepackage{bm}
\usepackage{amsmath,mathtools}
\usepackage{amssymb}
\usepackage[unicode=true,
 bookmarks=false,
 breaklinks=false,pdfborder={0 0 1},colorlinks=false]
 {hyperref}
\hypersetup{
colorlinks,citecolor=blue,filecolor=blue,linkcolor=blue,urlcolor=blue}

\makeatletter
\usepackage{amsthm}
\usepackage{comment}
\usepackage[round]{natbib}
\usepackage{booktabs}

\usepackage{graphicx}

\usepackage[linesnumbered,ruled,vlined]{algorithm2e}

\SetCommentSty{mycommfont}
\usepackage{algorithmic}

\usepackage[final]{showlabels}

\usepackage{float}
\usepackage{multirow}
 
\usepackage{dsfont}
\usepackage{tcolorbox}

\usepackage{color}
\definecolor{yxc}{RGB}{255,0,0}
\definecolor{yjc}{RGB}{125,0,0}
\definecolor{ytw}{RGB}{255,69,0}
\definecolor{gen}{RGB}{0,0,200}

\allowdisplaybreaks

\newcommand{\Var}{\mathsf{Var}}

\newcommand{\mymid}{\,|\,}

\newcommand{\Psource}{P_{\mathsf{base}}}
\newcommand{\Vsource}{\mathsf{TotVar}_{\mathsf{base}}}
\newcommand{\psource}{p_{\mathsf{base}}}
\newcommand{\Pdata}{P_{\mathsf{data}}}

\newcommand{\diff}{\mathrm{d}}

\definecolor{yanxi}{RGB}{0,200,100}

\title{Flow Matching from Non-Gaussian Sources: \\ A Theory of Efficient Sampling\footnotetext{The first two authors contributed equally.}}

\author{%
  Yuchen Zhou%
  \thanks{Department of Statistics, University of Illinois Urbana-Champaign.}
  \and
  Lili Zheng%
  \footnotemark[1]%
  \and
  Yuxin Chen%
  \thanks{Department of Statistics and Data Science, the Wharton School, University of Pennsylvania.}
  \and
  Gen Li%
  \thanks{Department of Statistics and Data Science, Chinese University of Hong Kong.}
}

\makeatother

\begin{document}

\theoremstyle{plain} \newtheorem{lemma}{\textbf{Lemma}}\newtheorem{proposition}{\textbf{Proposition}}\newtheorem{theorem}{\textbf{Theorem}}
\newtheorem{corollary}{\textbf{Corollary}}
\theoremstyle{plain}\newtheorem{assumption}{\textbf{Assumption}}
\theoremstyle{remark}\newtheorem{remark}{\textbf{Remark}}

\maketitle 

\begin{abstract}
Flow matching is a versatile framework for generative modeling that transports a source distribution to a target data distribution. A key advantage is its flexibility in the choice of source distribution, such as heavy-tailed and multimodal sources. Yet finite-step sampling guarantees for non-Gaussian sources remain limited. We establish a nonasymptotic convergence theory for Euler discretization of flow matching with linear interpolation. Suppose the source distribution on $\mathbb{R}^d$ has an everywhere positive, continuously differentiable density, finite total variance $\Vsource$, and an $L$-Lipschitz score. Using the population velocity field and a suitable time grid, this sampler achieves total variation error at most $\varepsilon$ in
\[
\widetilde{O}\!\left(\frac{Ld\Vsource}{\varepsilon}\right)
\]
iterations, provided the target distribution satisfies a mild second-moment condition. These assumptions cover, among many examples, source distributions such as multivariate Student's \(t\) distributions with more than two degrees of freedom and finite Gaussian mixtures with shared covariance, while accommodating target distributions with neither a smooth density nor bounded support. 
\end{abstract}



\section{Introduction}\label{sec:intro}

 Flow matching~\citep{lipman2023flow,albergo2022building,liu2023flow} has established itself as a powerful and versatile framework for generative modeling, achieving compelling performance in image generation~\citep{esser2024scaling},
speech synthesis~\citep{le2023voicebox}, and even molecule generation~\citep{song2023equivariant} and protein structure generation~\citep{bose2024se}. In a nutshell, flow matching seeks to transport a source distribution toward a target data distribution by means of an ordinary differential equation (ODE):
 \begin{equation}\label{eq:ODE}
	 \frac{\mathrm{d}}{\mathrm{d} t} X_t=v_t(X_t),\quad t\in [0,1),\qquad X_0\sim \Psource,
 \end{equation}
where $\Psource$ represents a source distribution on $\mathbb{R}^d$ that is easy to sample from, and \(v_t:\mathbb{R}^d \to \mathbb{R}^d\) stands for a time-dependent vector field (often referred to as the velocity field).  Ideally, the distribution of $X_t$ converges to the target data distribution --- denoted by $\Pdata$ --- as $t$ approaches $1$. Flow matching is typically composed of two stages: during training, the population vector field $v_t(\cdot)$ is learned by regression  using samples drawn from the source and target distributions; during sampling, one draws $X_0\sim \Psource$ and numerically solves the ODE~\eqref{eq:ODE} from this initialization $X_0$, using a standard approach such as Euler's method, to obtain an approximate sample from \(\Pdata\).

The popularity of flow matching stems in part from its simulation-free training, empirical effectiveness,  conceptual simplicity, and flexibility in choosing the source distribution and, in some formulations, the source-target coupling~\citep{liu2023flow,Tong2023ImprovingAG,pooladian2023multisample}. Given the widespread adoption of flow matching in generative modeling, an important practical consideration is its sampling efficiency. As with ODE-based samplers for diffusion models, generating a sample with flow matching requires numerically solving an ODE driven by a learned velocity field, typically by discretizing time into a grid and evaluating the field at each step. A finer time grid can reduce discretization error and improve sampling accuracy, but it also requires more evaluations of the velocity field, each typically involving a forward pass through a large neural network. To understand this tradeoff and guide the choice of time grid in practice, the present paper seeks to characterize the iteration complexity of the flow matching sampler, namely, the number of steps needed to achieve a prescribed sampling accuracy.

Notably, while substantial work has been devoted to establishing convergence theory for diffusion models ---  whose probability flow ODEs can be formulated as flow matching with Gaussian sources~\citep{albergo2025stochastic} --- the analyses and insights therein do not directly carry over to flow matching with non-Gaussian sources. This theoretical gap matters in practice because non-Gaussian sources can be highly versatile: they support sampling from heavy-tailed or multimodal distributions, data-to-data transport, and the generation of non-Euclidean data.  This motivates a closer examination of the role of non-Gaussian sources and the theoretical support for them.

 \subsection{Non-Gaussian sources in flow matching}

Conventionally, Gaussian sources are the standard choice in diffusion models: the forward process of a diffusion model gradually adds Gaussian noise until the data distribution approaches a Gaussian source, while preserving tractable forward transition distributions~\citep{Ho2020Denoising,song2021denoising,Song2021Scorebased}.   However, Gaussian sources may be less well suited to target distributions with heavy tails or to data with discrete or geometric structure.  Alternative source distributions have therefore been explored, with tractability as a key consideration; examples include Gamma and $\alpha$-stable noise with suitable closure properties under addition~\citep{Nachmani2021NonGD,NEURIPS2023_8011b23e,popov2025improved,Shariatian2024HeavyTailedDW}, and discrete corruption schemes that allow multi-step transition probabilities to be computed tractably~\citep{NEURIPS2021_958c5305}.

By contrast, flow matching offers greater flexibility in the choice of source distribution. Given samples from any source and target distributions, the velocity field can be learned through regression.  For heavy-tailed targets, prior work explored multivariate \(t\) distributions as sources, reporting improved performance over Gaussian-source baselines, while other work proposed to learn quantile-based source distributions that adapt to heavy-tailed data~\citep{Pandey2024HeavyTailedDM,guan2026mirror,chemseddineadapting}. Gaussian mixture sources have also been used effectively to generate multimodal targets~\citep{luo2026flow,wang2026meanfuser}. Other work employed non-Euclidean source distributions to generate discrete data or data lying on a manifold, such as DNA sequences, proteins, crystals, and geographic locations~\citep{Strk2024DirichletFM,Chen2023FlowMO,Miller2024FlowMMGM,bose2024se,sriram2024flowllm}. In addition, flow matching has been used for data-to-data transport, such as translating images between domains, where the source is itself a non-Gaussian data distribution~\citep{liu2023flow,Tong2023ImprovingAG}. Despite these promising applications of non-Gaussian sources, the effects of source choice on training and sampling remain under debate~\citep{cherkaoui2026heavy,lee2025there}. It is largely unclear to what extent existing convergence results for standard Gaussian sources carry over to more general sources.

A few recent studies began to analyze the convergence behavior of flow matching and its variants with non-Gaussian sources. Some assumed the learned velocity field is bounded~\citep{guan2026total}, an assumption that may be restrictive for source distributions with unbounded support. Others imposed tail conditions on the target data distribution~\citep{guan2026mirror} or introduced Gaussian smoothing along the flow trajectory through stochastic interpolants~\citep{liu2025finitetime,liu2026finite,silveri2026diffusion}. This motivates the following question:
\begin{quote}
{\em For commonly used non-Gaussian sources, can we establish nonasymptotic convergence guarantees without introducing Gaussian smoothing or imposing stronger assumptions on the target?}
\end{quote}
%


\subsection{This paper}

\paragraph{Main contributions.} 
In this paper, we address the above question by analyzing the Euler sampler for flow matching when the true velocity field is available. Our results hold under the following conditions on the source and target distributions, which are satisfied by a broad class of commonly used distributions:
\begin{itemize}
	\item The source distribution $\Psource$ has an everywhere positive and continuously differentiable density on $\mathbb{R}^d$, finite total variance, and an $L$-Lipschitz score function.
    
	\item The target distribution $\Pdata$ has a finite second moment that is not overly large relative to the source's second moment.
\end{itemize}
Under these conditions, we prove that the Euler sampler for flow matching achieves a total-variation-based error of at most $\varepsilon$ using at most
\begin{equation}
\widetilde{O}\Bigg(\frac{Ld \,\Vsource}{\varepsilon}\Bigg)
\end{equation}
evaluations of the velocity field, 
where $\Vsource \coloneqq \mathop{\mathbb E}\nolimits_{Z\sim \Psource}\left[\|Z-\mathbb E[Z]\|_2^2\right]$ represents the total variance of the source distribution.


Our convergence theory is established under substantially milder assumptions on the source and target distributions than those imposed in prior work. For the source distribution, the assumptions of a Lipschitz score and a finite second moment are satisfied, for example, by multivariate Gaussian distributions with positive definite covariance matrices, multivariate Student's $t$ distributions with more than two degrees of freedom, and finite Gaussian mixtures whose components share a positive definite covariance matrix; note that both Student's $t$ distributions and Gaussian mixtures have been used as non-Gaussian sources in flow matching applications~\citep{Pandey2024HeavyTailedDM,luo2026flow}. For the target distribution, we require only a finite second moment that is not too large relative to that of the source, with no additional tail assumptions. These results demonstrate that flow matching samplers can efficiently transport a broad class of non-Gaussian source distributions, including heavy-tailed distributions, to fairly general target distributions.

From a technical standpoint, our analysis rests on carefully characterizing the nonlinear score of the source distribution and establishing a restricted injectivity property for the Euler map. Both of these technical steps are more involved than in the Gaussian case, the latter of which features an affine source score and a globally injective Euler map.


\paragraph{Organization.} 
The remainder of this paper is organized as follows. Section~\ref{sec:prelim} formally introduces the flow matching sampler.  Section~\ref{sec:main} presents our main convergence theory and its consequences, followed by additional related work in Section~\ref{sec:related-work}.  Section~\ref{sec:proof_highlight} outlines the proof strategy and highlights the key steps for handling non-Gaussian source distributions.  Section~\ref{sec:analysis} proves the main theorem, with proofs of auxiliary results deferred to the appendix.  Section~\ref{sec:discussion} concludes the paper by discussing directions for future work.

\paragraph{Notation.}
We use the following notational conventions throughout. 
Capital letters $X_\star,Z,X_t,Y_n$ denote random vectors, while lowercase letters $x_\star,z,x,y_n$ denote fixed values or integration variables. For any random vector \(W\), let \(P_W\) denote its law and \(p_W\) its Lebesgue density when it exists. We write \(p_{X_\star\mid X_t}(\cdot\mid x)\) for the density of the conditional law of \(X_\star\) given \(X_t=x\), with respect to \(\Pdata = P_{X_\star}\). For probability laws $P$ and $Q$, define their total-variation (TV) distance as
$$\mathsf{TV}(P,Q)=\frac12\int|\mathrm dP-\mathrm dQ|.$$
We also adopt the Jacobian convention
$[\nabla_xu]_{ij}=\partial u_i/\partial x_j$. For a symmetric matrix $\Sigma$, $\lambda_{\min}(\Sigma)$ denotes the smallest eigenvalue of $\Sigma$.

We use $f(\cdot)=O(g(\cdot))$ to mean that $f\le C_0g$ for some universal constant $C_0>0$. The notation $\widetilde{O}(\cdot)$ has the same meaning as $O(\cdot)$, except that it hides the logarithmic factors. Unless stated otherwise, $C,c>0$ denote numerical constants whose values may change from line to line.

\section{Preliminaries}\label{sec:prelim}

In this section, we formalize the flow-matching setup and the Euler sampler studied in this paper. 

\paragraph{Flow matching setup.} 
Recall that  $\Psource$ and $\Pdata$ denote the source and target distributions on $\mathbb R^d$, respectively. Flow matching models the transport from a source distribution \(\Psource\) to \(\Pdata\) through an ODE driven by a time-dependent velocity field:
\begin{align}\label{eq:CNF}
	\frac{\mathrm{d}X_t}{\mathrm{d}t}=v_t(X_t), \qquad X_0\sim \Psource,
\end{align}
where the velocity field $\{v_t(\cdot)\}_{t\in[0,1)}$ is assumed to be sufficiently regular to ensure existence and uniqueness of solutions to this ODE. Let $\phi_t:\mathbb{R}^d\to\mathbb{R}^d$ denote the corresponding flow map, which maps each initial condition at time $0$ to the solution at time $t$, i.e., 
\begin{align}
    X_t=\phi_t(X_0).
\end{align}
Throughout this paper, we let $P_t=P_{X_t}$  denote the law of $X_t$ for $0\le t<1$. We assume that $\Psource$ and each $P_t$ ($0\le t<1$) admit densities $\psource$ and $p_t$, respectively; in particular, $p_0=\psource$. For an ideal velocity field, $P_t$ converges weakly to $\Pdata$ as $t\uparrow 1$.

Flow matching trains a continuous normalizing flow by regressing a parameterized velocity field onto a target field that generates a probability path $\{P_t\}_{t\in[0,1]}$ connecting the source and target distributions. A core idea behind flow matching is to construct this path using conditional flows, which we introduce next.

\paragraph{Conditional flows.}
A useful way to construct a probability path from $\Psource$ to $\Pdata$ is to first specify how a source sample moves toward a given target sample, then average over target samples. More precisely, let $X_\star\sim\Pdata$ and $Z\sim \Psource$ be independent samples from the target and source distributions. Conditional on $X_\star=x_\star$, consider the flow defined by
\begin{align}\label{eq:conditional-velocity}
	\frac{\mathrm{d}X_t}{\mathrm{d}t}=v_t(X_t\mid x_\star),\quad X_0=Z,
	\qquad
	v_t(x\mid x_\star)=\frac{x_\star-x}{1-t},
	\quad t\in[0,1).
\end{align}
In words, $v_t(\cdot\mid x_\star)$ is the conditional velocity field that points from the current state toward the fixed target $x_\star$. Solving this ODE~\eqref{eq:conditional-velocity} for each target value yields the linear interpolant
\begin{equation}
	X_t=(1-t)Z+tx_\star,
	\qquad t\in[0,1),
	\label{eq:linear-interpolant-xstar}
\end{equation}
where $X_1=x_\star$ by continuous extension. For $t<1$, the resulting conditional density obeys
\begin{align}\label{eq:conditional-density}
	p_{X_t\mid X_\star}(x\mid x_\star)
	=\frac{1}{(1-t)^d}
	\psource\bigg(\frac{x-tx_\star}{1-t}\bigg).
\end{align}
This straight-line interpolation corresponds to the conditional optimal transport path in~\citet{lipman2023flow};  it was also used by~\citet{liu2023flow} to construct 1-rectified flow.

\paragraph{Marginal flow.}
Marginalizing the conditional densities over $X_\star\sim \Pdata$ yields the marginal probability path
\begin{align}
	p_t(x) \coloneqq p_{X_t}(x)
	=\int_{\mathbb R^d}p_{X_t\mid X_\star}(x\mid x_\star)\,\Pdata(\diff x_\star),
	\qquad t\in[0,1),
\end{align}
with $p_0=\psource$, and $P_t$ converges weakly to $\Pdata$ as $t\uparrow 1$.  The corresponding marginal velocity field satisfies
\begin{align}
	v_t(x)
	=\mathbb E[v_t(x\mid X_\star)\mid X_t=x]=\mathbb E[X_\star-Z\mid X_t=x]=\frac{\mathbb E[X_\star\mid X_t=x]-x}{1-t}. 
	\label{eq:marginal-velocity}
\end{align}
Here, the conditional expectation is taken under the joint law of $(X_t,X_\star)$ corresponding to
\begin{equation}
	X_t=(1-t)Z+tX_\star,
	\qquad Z\sim \Psource, \qquad X_{\star}\sim \Pdata,   
	\label{eq:linear-interpolant}
\end{equation}
with $Z$ and $X_\star$ independent. 
Under suitable integrability and regularity assumptions, the marginal density and velocity field satisfy the continuity equation or transport equation~\citep{lipman2023flow}:
\[
\partial_t p_t+\nabla_x\cdot(p_tv_t)=0.
\]
Consequently, under these regularity conditions, the solution of ODE~\eqref{eq:CNF} with $v_t(\cdot)$ defined in \eqref{eq:marginal-velocity} satisfies $X_t\sim P_t$ at each time $t$.


\paragraph{Euler discretization for sampling.}
To generate samples in practice, the continuous-time ODE must be approximated numerically. We focus on the Euler sampler --- a standard, first-order discretization of the continuous-time ODE --- with a rescaled initialization and early stopping. Given a time grid $0<t_0<t_1<\cdots<t_N<1$ of $N+1$ time points, the Euler method proceeds as
\begin{align}
	Y_0 \overset{\mathrm d}{=} (1-t_0)Z,\qquad
	Y_n = Y_{n-1} + v_{t_{n-1}}(Y_{n-1})(t_n-t_{n-1}),
	\quad n=1,\ldots,N.
	\label{eq:sampler}
\end{align}
At each step, the velocity field is evaluated at the left endpoint \(t_{n-1}\), and the sampler outputs \(Y_N\).  As will be specified in Section~\ref{sec:assumptions}, both $t_0$ and $1-t_N$ are chosen to be sufficiently small. 
The positive initial time \(t_0\) is a technical modification that requires only rescaling the initial noise. Early stopping at $t_N<1$ is common in the literature~\citep{Chen2023Improved,chewi2025ddpm}. As we will shortly see, this allows us to establish total-variation-based guarantees relative to the intermediate law \(P_{t_N}\) (which has a continuously differentiable density $p_{t_N}$), since the data distribution \(\Pdata\) may be singular.
\begin{remark}
    In this paper, we focus on the sampling error and therefore consider the Euler sampler with the population velocity field $v_t$, as defined in \eqref{eq:sampler}. In practice, one often uses a learned velocity field $\widehat{v}_t$, trained by regressing $X_\star-Z$ upon $X_t = (1-t)Z+tX_\star$. 
\end{remark}

\section{Main results}\label{sec:main}

In this section, we establish convergence guarantees for the Euler method for flow matching,  by bounding the total-variation distance between the sampler's output distribution $P_{Y_N}$ and the law \(P_{t_N}=P_{X_{t_N}}\) of the interpolant at time \(t_N\). Note that the law $P_{t_N}$ has a continuously differentiable density and converges weakly to the data distribution \(\Pdata\) as \(t_N\uparrow 1\).

\subsection{Assumptions}
\label{sec:assumptions}

We begin by stating the assumptions needed for our theoretical analysis, where the first assumption is concerned with the source distribution.
\begin{assumption}\label{ass:source}
	The source distribution $\Psource$ has a strictly positive, continuously differentiable (\(C^1\)) density \(\psource\) on \(\mathbb R^d\), 
	and obeys $\mathbb E_{Z\sim \Psource}\left[\|Z-\mathbb E[Z]\|_2^2\right]<\infty$. Additionally, its score function $s(\cdot)=\nabla\log \psource(\cdot)$ is \(L\)-Lipschitz, namely,
	\begin{equation}
		\big\|s(z)-s(z')\big\|_2\leq L\|z-z'\|_2
		\qquad \text{for all }z,z'\in\mathbb R^d.
		\label{eq:source-lipschitz}
	\end{equation}
\end{assumption}
This assumption concerns only the source distribution, which can be chosen by the practitioner. It is satisfied, for example, by multivariate Gaussian distributions with positive-definite covariance matrices, multivariate Student's \(t\) distributions with more than two degrees of freedom, and finite Gaussian mixtures whose components share a positive-definite covariance matrix. Student's \(t\) distributions and Gaussian mixtures have both been used as non-Gaussian sources in flow-matching applications~\citep{Pandey2024HeavyTailedDM,luo2026flow}. For a new source distribution, the assumption can be checked by verifying the regularity of its density and score.

The next assumption relates the target's second moment to the source's centered second moment.
\begin{assumption}\label{ass:target}
Let \(X_\star\sim\Pdata\) and \(Z\sim\Psource\). 
	Assume that there exists a fixed numerical constant $c_R>0$ such that
	\begin{equation}
		\mathbb E\left[\|X_\star\|_2^2\right]
		\leq (N+1)^{c_R}\mathbb E\left[\|Z-\mathbb E[Z]\|_2^2\right].
		\label{eq:target-moment}
	\end{equation}
\end{assumption}
Assumption~\ref{ass:target} allows the target's second moment to grow polynomially with the number $N$ of sampling steps, while controlling it relative to the source's centered second moment. In particular, it favors choosing a source that is not too concentrated relative to the target.

Last but not least, we impose a condition on the discretization time grid. It requires the initial and stopping times to lie near the endpoints and controls the step sizes relative to their distance from those endpoints.
\begin{assumption}\label{ass:schedule}
	Fix constants $c_0\geq c_R+4$ and $c_s>0$, and suppose
	$c_s\log(N+1)/(N+1)\leq1/2$. The sampling times
	$0<t_0<t_1<\cdots<t_N<1$ satisfy
	\begin{equation}
		\begin{split}
			t_0&\leq(N+1)^{-c_0/2},\qquad
			1-t_N\leq(N+1)^{-c_0/2},\\
			\frac{t_n-t_{n-1}}{t_{n-1}(1-t_n)}
			&\leq\frac{c_s\log(N+1)}{N+1},\qquad 1\leq n\leq N.
		\end{split}
		\label{eq:time-grid-condition}
	\end{equation}
\end{assumption}

For concreteness, we note that 
Assumption~\ref{ass:schedule} can be met by, for example, either of the following schedules:
\begin{itemize}
	\item[(i)] $t_n=\left(1+(N+1)^{c_0(1/2-n/N)}\right)^{-1}$.
    
	\item[(ii)] The times increase geometrically near zero, and approach one geometrically near the end:
	\[
	t_n=
	\begin{cases}
		\varepsilon q^n, & 0\le n\le \lfloor N/2\rfloor,\\
		1-\varepsilon q^{N-n}, & \lfloor N/2\rfloor< n\le N,
	\end{cases}
	\]
	where
	$\varepsilon=(N+1)^{-c_0/2}$ and $q=(2\varepsilon)^{-2/N}$. 
\end{itemize}
Both schedules use smaller steps near the endpoints. Related grids also appear in convergence analyses of diffusion models~\citep{jiao2025optimal}.

\subsection{Convergence theory and implications}
\label{eq:theory-implications}

We are now positioned to state the main convergence guarantee. For notational convenience, we use
\begin{align}
\Vsource
\coloneqq 
\mathop{\mathbb E}\limits_{Z\sim\Psource}\big[\|Z-\mathbb E[Z]\|_2^2\big]
\end{align}
to denote the total variance of the source distribution $\Psource$ throughout the rest of the paper. 
\begin{theorem}[Sources with a finite second moment]\label{thm:general-moments}
	Suppose $N\geq2$, and Assumptions~\ref{ass:source}, \ref{ass:target} and
	\ref{ass:schedule} hold. Then the Euler sampler for flow matching (cf.~\eqref{eq:sampler}) satisfies
	\begin{equation}
		\mathsf{TV}(P_{Y_N},P_{t_N})
		\leq 
		\frac{CL(d+\log N)\Vsource\log^2N}{N}
		\label{eq:general-second-final}
	\end{equation}
    for some universal constant $C>0$. 
\end{theorem}

Theorem~\ref{thm:general-moments} develops an upper bound on the TV distance between the Euler sampler's output distribution $P_{Y_N}$ and the true marginal law $P_{t_N}=P_{X_{t_N}}$. By Assumption~\ref{ass:schedule}, the sampler is terminated at a time $t_N$ satisfying $1-(N+1)^{-c_0/2}\leq t_N<1$. Note that  $P_{t_N}$ is a slightly perturbed version of the target distribution, representing the law of $$t_NX_\star+(1-t_N)Z\sim P_{t_N}$$ with $(X_\star,Z)\sim \Pdata\times \Psource$. As alluded to previously, we compare the sampler with this intermediate law because the target distribution \(\Pdata\) may be singular and need not have a density, whereas \(P_{t_N}\) has a continuously differentiable density. This way of comparison is also common in convergence analyses of diffusion models~\citep{Chen2023Improved,chewi2025ddpm}.
Furthermore, it is noteworthy that Theorem~\ref{thm:general-moments} is established under minimal assumptions on the source and target distributions: a Lipschitz score condition on the source, a mild second-moment bound on the target (relative to the source), and a standard stepsize condition. This confirms the flow matching sampler's capability in efficiently transporting fairly general source distributions to target distributions. 


A few additional remarks are as follows. 
\begin{itemize}
\item {\em Comparison with prior convergence theory for non-Gaussian sources.} 
	Perhaps the most closely related results to ours are those of \citet{guan2026mirror,guan2026total}. \citet{guan2026mirror} established Wasserstein bounds for flow matching in the dual space with a Student-$t_\nu$ source, encompassing Euclidean flow matching as a special case. However, its guarantee requires a bounded target density in dual coordinates satisfying $p_1(x)\lesssim \|x\|^{-2d-\nu-2}$ for $\|x\|\ge1$. This condition excludes discrete targets and target densities with heavier tails. \citet{guan2026total} established TV bounds for Riemannian flow matching, with Euclidean flow matching covered by its result for Hadamard manifolds. They imposed additional regularity assumptions on the velocity fields, including boundedness assumptions. Notably, the boundedness condition may fail to hold even for a standard Gaussian source.

\item {\em Consequences of Theorem~\ref{thm:general-moments} for commonly used source distributions.} 
	The source distribution mainly affects the TV-based guarantee through the factor $L\,\Vsource$. Here,  we evaluate this quantity analytically for a few examples; the calculations are postponed to Appendix~\ref{app:source-examples}.
	\begin{itemize}
		\item {\em Gaussian source.} Let \(Z\sim\mathcal N(\mu,\Sigma)\) with mean \(\mu\in\mathbb R^d\) and covariance matrix
    \(\Sigma\succ0\). Then
        \begin{align}L\,\Vsource = \frac{{\rm tr}(\Sigma)}{\lambda_{\min}(\Sigma)}.
        \end{align}
        
		\item {\em Multivariate Student's $t$ source.} 
        Let \(Z\sim t_\nu(\mu,\Sigma)\) have \(\nu>2\) degrees of freedom,
    location \(\mu\in\mathbb R^d\), and scale matrix
    \(\Sigma\succ 0\). In this case, one has
    \begin{align}L\,\Vsource=\frac{\nu+d}{\nu-2}\frac{{\rm tr}(\Sigma)}{\lambda_{\min}(\Sigma)}.
        \end{align}
        
		\item {\em Gaussian mixture.}  Let
    \[
    Z\sim\sum_{k=1}^{m}\pi_k\mathcal N(\mu_k,\Sigma),
    \]
    where \(\pi_k>0\), \(\sum_{k=1}^{m}\pi_k=1\),
    \(\mu_k\in\mathbb R^d\), and \(\Sigma\succ0\). Then one has
        \begin{align}
        L\, \Vsource =\frac{\operatorname{tr}(\Sigma)+B}
		{\lambda_{\min}(\Sigma)}
		\max\left\{1,\frac{\Delta^2}{4}-1\right\},
        \end{align}
        where we take $\overline\mu=\sum_k\pi_k\mu_k$ and \[B=\sum_{k=1}^m\pi_k\|\mu_k-\overline{\mu}\|_2^2,\qquad\Delta^2=\max_{1\leq i,j\leq m}(\mu_i-\mu_j)^\top \Sigma^{-1}(\mu_i-\mu_j).\] 
	\end{itemize}
	In each of the above examples, $L\,\Vsource$ depends on $\lambda_{\min}(\Sigma)^{-1}$ together with a measure of total spread, e.g., $\mathrm{tr}(\Sigma)$. 
    This suggests the potential benefit of a well-conditioned source: a source with better-conditioned covariance  may facilitate faster sampling for a broad class of target distributions.


\end{itemize}

\section{Related work}
\label{sec:related-work}

The flow matching framework was first introduced by \citet{lipman2023flow} as a simulation-free approach to learning velocity fields for a range of probability paths, including diffusion paths and conditional optimal transport (OT) paths. In concurrent work, \citet{liu2023flow} introduced rectified flow using straight-line interpolation between source and target samples, which coincides with the interpolation rule that \citet{lipman2023flow} used in their conditional OT path when the terminal noise is zero. \citet{liu2023flow} also proposed reflow, using endpoint pairs from a previously learned flow to train a new flow with straighter trajectories. \citet{pooladian2023multisample} and \citet{Tong2023ImprovingAG} pursued a similar goal by coupling source and target samples within each minibatch. In complementary work, \citet{albergo2022building} introduced a stochastic interpolant framework that supports both linear and trigonometric interpolation; \citet{albergo2025stochastic} later extended it to allow auxiliary noise in the interpolant. In the current paper, we focus on the flow induced by straight-line interpolation between independent source and target samples, which is also the $1$-rectified flow of \citet{liu2023flow}.

Our work is intimately related to prior convergence theory developed for both flow matching and diffusion models, which often considers two types of error. The first is the statistical error in estimating the velocity field or score function from a finite training sample~\citep{oko2023diffusion,fukumizu2025flow,chewi2025ddpm}. The second is the sampling error, which concerns how discretization affects the accuracy of the sampler. We focus on the sampling error in this work and assume that the true velocity field is available. For diffusion models, there is a rich literature on sampling convergence under Gaussian corruption, including results for the deterministic DDIM sampler~\citep{song2021denoising,Chen2023Restoration,li2024sharp,liang2025low}, the stochastic DDPM sampler~\citep{Ho2020Denoising,Chen2023Improved,li2024adapting,li2025d,azangulov2024convergence,potaptchik2024linear,huang2026denoising,jiao2025optimal}, and their higher-order variants~\citep{li2024accelerating,wu2024stochastic,li2025faster,jiao2026forward}. The connection between rectified flow and DDPM was leveraged to study low-dimensional adaptation of rectified flow in \citet{roy2026low} under Gaussian sources.  A few studies also develop convergence theory for specific non-Gaussian corruptions, such as $\alpha$-stable noise~\citep{hammani2026global}. 

A strand of recent work has studied convergence properties for flow matching and its variants (e.g., \citet{liu2025finitetime,liu2026finite,silveri2026diffusion,benton2024error,cheng2024convergence,bansal2025wasserstein,roy2026low,guan2026mirror,guan2026total,wang2026computational,chambolle2026limit,xie2026flow,pham2026setting}).  For instance, \citet{liu2025finitetime,liu2026finite,silveri2026diffusion} studied ODE or SDE samplers for stochastic interpolants that introduce Gaussian noise along the flow trajectory. \citet{guan2026mirror} established Wasserstein bounds for mirror flow matching with a Student's $t$ source under a polynomial tail condition on the target density. \citet{guan2026total} established total variation bounds for Riemannian flow matching under regularity and boundedness assumptions on the population and learned velocity fields. In comparison, the present paper establishes a total variation bound for the Euler sampler with linear interpolation and no auxiliary Gaussian noise. Our result applies to a broad class of non-Gaussian sources and requires neither a bounded velocity field nor a polynomial tail bound on the target density.


\section{Technical novelty}\label{sec:proof_highlight}
Our theory accommodates general non-Gaussian sources, including heavy-tailed sources such as the multivariate Student's $t$ distributions. Accordingly, we do not assume exponential tails or higher moments beyond the second.  This creates two major obstacles in our analysis: the one-step Euler map does not inherit the injectivity of Gaussian source, and the local density-comparison error analysis cannot assume arbitrarily high-order moments to be finite; we also need to deal with multiple remainder terms due to non-Gaussian source throughout the proof. In this section, we briefly describe our main proof framework and highlight our strategies for tackling non-Gaussian source-related challenges.

\paragraph{Decomposition of the final TV error.}
First of all, different from prior analysis for DDIM convergence~\citep{li2024sharp} where density ratio error at each step is accumulated across the whole sampling trajectory, we make the following observation:
\[
\mathsf{TV}(P_{Y_N},P_{t_N})
\le
\mathsf{TV}(P_{Y_0},P_{t_0})
+\sum_{n=1}^N
\mathsf{TV}(P_{\widehat Y_n},P_{t_n}),
\]
where $\widehat Y_n=T_n(X_{t_{n-1}})$ and $T_n$ is the one-step Euler update function:
\[
T_n(X_{t_{n-1}})= X_{t_{n-1}}+(t_n-t_{n-1})v_{t_{n-1}}(X_{t_{n-1}}).
\]
 $P_{t_n}$ is the distribution of $X_{t_n}=X_{t_{n-1}} + (t_n-t_{n-1})(X_\star-Z)$. Thus, each summand compares one Euler update from the exact marginal $P_{t_{n-1}}$ with the next marginal.

When the map $T_n$ is $C^1$ and globally injective with image space $\mathbb{R}^d$, then for every Borel set $A$,
\begin{align*}
	\mathbb{P}(\hat{Y}_n\in A)&= \int_{T_n^{-1}(A)}p_{t_{n-1}}(y)\,{\rm d}y,\\
	\mathbb{P}(X_{t_n}\in A)&= \int_{T_n^{-1}(A)}p_{t_n}(T_n(y))|\det\nabla T_n(y)|\,{\rm d}y.
\end{align*}
Hence, 
\[
\mathsf{TV}(P_{\widehat Y_n},P_{t_n})
=\int p_{t_{n-1}}(x)
\left[1-\frac{p_{t_n}(T_n(x))|\det\nabla T_n(x)|}{p_{t_{n-1}}(x)}\right]_+\,{\rm d}x.
\]
Consequently, a lower bound on the combined log-density and log-determinant ratio suffices. However, the global injectivity of $T_n$ follows for a standard Gaussian source but may not hold for more general sources. In fact,
\begin{equation}\label{eq:nabla_T}
	\begin{aligned}
		\nabla T_n(x)
		&=\frac{1-t_n}{1-t_{n-1}}I_d-\frac{(t_n-t_{n-1})}{t_{n-1}(1-t_{n-1})}\operatorname{Cov}(Z,s(Z)\mid X_{t_{n-1}}=x)\\
		&\overset{Z\sim\mathcal N(0,I_d)}{=}
		\frac{1-t_n}{1-t_{n-1}}I_d
		+\frac{(t_n-t_{n-1})}{t_{n-1}(1-t_{n-1})}\operatorname{Cov}(Z\mid X_{t_{n-1}}=x)
		\succeq\frac{1-t_n}{1-t_{n-1}}I_d,
	\end{aligned}
\end{equation}
The second equality uses the standard Gaussian score $s(z)=-z$. The uniform positive lower bound on $\nabla T_n$ gives strong monotonicity and hence global injectivity. For a general source, $-\operatorname{Cov}(Z,s(Z)\mid X_{t_{n-1}}=x)$ need not be symmetric positive semidefinite, so the same argument does not apply. We therefore construct a Borel set $G_n\subseteq\mathbb R^d$ on which $T_n$ is injective and has positive determinant. The localized comparison at each step $t_n$ then takes the following form:
\begin{equation}\label{eq:highlight-restricted-tv}
	\begin{aligned}
		\mathsf{TV}(P_{\widehat Y_n},P_{t_n})
		&\leq\mathbb P(X_{t_{n-1}}\notin G_n)+\int_{G_n}p_{t_{n-1}}(x)
		\left[1-\frac{p_{t_n}(T_n(x))\det\nabla T_n(x)}{p_{t_{n-1}}(x)}\right]_+\,{\rm d}x.
	\end{aligned}
\end{equation}

\paragraph{Lower bounds for the density ratio and Jacobian determinant.}
For $x\in G_n$, we need a lower bound close to zero for $\log[p_{t_n}(T_n(x))/p_{t_{n-1}}(x)]+\log\det\nabla T_n(x)$. One main obstacle is that the score function of the source distribution $s(Z)$, a key quantity in both the density ratio term and the Jacobian determinant term, no longer takes the simple form $s(Z)=-Z$ as in the Gaussian case. Luckily, with the aid of the Lipschitz property of the score function, we obtain the following two lower bounds:
\begin{equation}\label{eq:density_ratio_jacobian_bnds}
	\begin{aligned}
		\log\frac{p_{t_n}(T_n(x))}{p_{t_{n-1}}(x)}
		&\geq -(t_n-t_{n-1})\operatorname{tr}\nabla v_{t_{n-1}}(x)\\
		&\quad-\frac{3L}{2}
		\left(\frac{t_n-t_{n-1}}{t_{n-1}(1-t_n)}\right)^2
		\mathbb E\!\left[
		\left\|Z-\mathbb E[Z\mid X_{t_{n-1}}=x]\right\|_2^2
		\,\middle|\,X_{t_{n-1}}=x\right],\\[6pt]
		\log\det\nabla T_n(x)
		&\geq +(t_n-t_{n-1})\operatorname{tr}\nabla v_{t_{n-1}}(x)-\frac43(t_n-t_{n-1})^2
		\|\nabla v_{t_{n-1}}(x)\|_{\rm F}^2.
	\end{aligned}
\end{equation}
The first-order contributions cancel, while the remaining cross-covariance coefficient is quadratic in step size $t_{n}-t_{n-1}$. To achieve the above, only the second moment of $Z$ and the Lipschitz property of the source score function are needed here. 
Summing the second-order terms in~\eqref{eq:density_ratio_jacobian_bnds} across $1\leq n\leq N$, we have a good control for $\sum_{n=1}^N\mathsf{TV}(P_{\widehat Y_n},P_{t_n})$ up to the sum of the discarded probabilities $\mathbb P(X_{t_{n-1}}\notin G_n)$.

\paragraph{Proof of restricted injectivity.}
Finally, we give a brief description over the construction of $G_n$ and the restricted injectivity proof, the key technical novelty of our proof. We also note that this may be of independent interest for future works dealing with non-Gaussian sources, where $T_n$ is likely not injective over the whole $\mathbb{R}^d$. In particular, motivated by \eqref{eq:nabla_T}, we construct $G_n$ such that the second term in \eqref{eq:nabla_T} is well-controlled to not ruin the positive definiteness of $\nabla T_n$:
\begin{align*}
	G_n=\Big\{x:\mathbb E[\|s(Z)\|_2^2\mid X_{t_{n-1}}=x]\leq A_1, \,\mathbb E[\|Z-\mathbb E[Z]\|_2^2\mid X_{t_{n-1}}=x]\leq A_2\Big\},
\end{align*}
where $A_1\asymp d$ and $A_2\gtrsim N^2/d$ up to logarithmic factors. Here, although $Z$ may be heavy-tailed, its score is sub-Gaussian, implied by the mild Lipschitz property of the score (see \cite{negrea2022approximations,altschuler2025shifted}; also Lemma~\ref{lem:source-estimates}). Combining sub-Gaussian tail control for $s(Z)$ and a Markovian tail control for $Z$ gives a good control for $\mathbb P(X_{t_{n-1}}\notin G_n)$.

Although our constructed good set $G_n$ ensures $\nabla T_n$ (symmetrized) to have lower bounded minimum eigenvalues, $T_n$'s injectivity on $G_n$ does not follow directly, as $G_n$ may not be a convex set. Instead, we design a new proof argument by showing that $T_n(x)=T_n(y)$ implies that the posterior mean of the source $Z$ given $X_{t_{n-1}}=x$ and $X_{t_{n-1}}=y$ must be sufficiently close, making it impossible for $x\neq y$ to happen. Appendix~\ref{app:restriction-proof} gives the argument for Lemma~\ref{lem:restriction}, the restricted injectivity statement. 

Finally, putting all pieces together and summing the TV errors over the grid $1\leq n\leq N$, we arrive at our final result:
\[
\mathsf{TV}(P_{Y_N},P_{t_N})
\lesssim L\Vsource (d+\log N)\frac{\log^2N}{N}.
\]

\section{Analysis}\label{sec:analysis}

In this section, we prove Theorem~\ref{thm:general-moments} by decomposing the total error into initialization and one-step discretization errors.

\paragraph{Preparation.}
The following lemma introduces two important properties for the sampling times and its proof is deferred to Section~\ref{sec:proof_grid}.
\begin{lemma}\label{lem:grid-estimates}
	Under Assumption~\ref{ass:schedule}, we have
	\begin{align}
		t_n-t_{n-1}&\leq\frac{c_s\log(N+1)}{N+1}
		t_{n-1}(1-t_{n-1}),\qquad 1\leq n\leq N,\notag\\
		\sum_{n=1}^N\left(\frac{t_n-t_{n-1}}{t_{n-1}(1-t_n)}\right)^r
		&\leq c_s^r\frac{\log^r(N+1)}{(N+1)^{r-1}},\qquad r\geq1.
		\label{eq:time-grid-sums}
	\end{align}
\end{lemma}

The next lemma provides some key properties of the score function, with its proof postponed to Section~\ref{app:source-moments}. The sub-Gaussian property of Lipschitz score functions has also been introduced in \cite{negrea2022approximations,altschuler2025shifted}.
\begin{lemma}[Score and second-moment estimates]\label{lem:source-estimates}
	Under Assumption~\ref{ass:source}, for every $v\in\mathbb R^d$,
	\begin{equation}
		\mathbb E\bigl[e^{v^\top s(Z)}\bigr]\leq e^{L\|v\|_2^2/2},\qquad
		\mathbb E[s(Z)]=0,\qquad
		\mathbb E[s(Z)s(Z)^\top]\preceq LI_d.
		\label{eq:source-score-mgf}
	\end{equation}
	Moreover, we have
	\begin{subequations}
		\begin{align}
			\mathbb E\bigl[e^{\|s(Z)\|_2^2/(4L)}\bigr]&\leq2^{d/2},
			\label{eq:source-score-norm-mgf}\\
			\mathbb E[(Z-\mathbb E[Z])s(Z)^\top]&=-I_d,
			\label{eq:source-Stein}\\
			L\mathbb E\bigl[\|Z-\mathbb E[Z]\|_2^2\bigr]&\geq d.
			\label{eq:scale-lower-bounds}
		\end{align}
	\end{subequations}
\end{lemma}

In addition, the lemma below shows that the initialization error can be well-controlled.
\begin{lemma}[Initialization error]\label{lem:initialization-absorbed}
	Suppose that Assumptions~\ref{ass:source}, \ref{ass:target}, and
	\ref{ass:schedule} hold. Then
	\begin{align}
		\mathsf{KL}(P_{t_0}\|P_{Y_0})
		&\leq\frac{Lt_0^2}{2(1-t_0)^2}\mathbb E\left[\|X_\star\|_2^2\right],
		\label{eq:init-KL}\\
		\mathsf{TV}(P_{Y_0},P_{t_0})
		&\leq\frac{t_0}{2(1-t_0)}
		\sqrt{L\mathbb E\left[\|X_\star\|_2^2\right]} \leq\frac{\sqrt{L\mathbb E\left[\|Z-\mathbb E[Z]\|_2^2\right]}}{(N+1)^2}.
		\label{eq:init-absorbed}
	\end{align}
\end{lemma}
The proof of Lemma~\ref{lem:initialization-absorbed} is given in Appendix~\ref{app:initialization}.

\paragraph{Step 1: Decomposing the total variation error.} 
We let
\begin{align}\label{def:hat_Y_n}
	\widehat Y_n=X_{t_{n-1}}+(t_n-t_{n-1})v_{t_{n-1}}(X_{t_{n-1}}) = T_n(X_{t_{n-1}}),\qquad
	\widehat P_n=P_{\widehat Y_n}.
\end{align}
Since total variation contracts under a measurable map, one has
\begin{align*}
	\mathsf{TV}(P_{Y_n},P_{t_n})
	&\leq\mathsf{TV}(P_{Y_n},\widehat P_n)
	+\mathsf{TV}(\widehat P_n,P_{t_n})\\*
	&\leq\mathsf{TV}(P_{Y_{n-1}},P_{t_{n-1}})
	+\mathsf{TV}(\widehat P_n,P_{t_n}).
\end{align*}
Repeatedly applying the previous inequality leads to
\begin{equation}
	\mathsf{TV}(P_{Y_N},P_{t_N})
	\leq\mathsf{TV}(P_{Y_0},P_{t_0})
	+\sum_{n=1}^N\mathsf{TV}(\widehat P_n,P_{t_n}).
	\label{eq:TV-telescope}
\end{equation}
To bound $\mathsf{TV}(P_{Y_N},P_{t_N})$, it suffices to control $\mathsf{TV}(\widehat P_n,P_{t_n})$ for each $n$.
We also define the following ``good'' event:
\begin{equation}
	\begin{split}
		G_n=\biggl\{x:\;&
		\mathbb E[\|s(Z)\|_2^2\mid X_{t_{n-1}}=x]
		\leq4L\{d+12\log(N+1)\},\\
		&\mathbb E[\|Z-\mathbb E[Z]\|_2^2\mid X_{t_{n-1}}=x]
		\leq\frac{t_{n-1}^2(1-t_n)^2}
		{256L(t_n-t_{n-1})^2\{d+12\log(N+1)\}}\biggr\}.
	\end{split}
	\label{eq:good-set}
\end{equation}
The following two lemmas give the regularity and injectivity needed
for the one-step comparison.

\begin{lemma}[Regularity needed for the one-step comparison]
	\label{lem:source-only-regularity}
	Under Assumption~\ref{ass:source}, for every fixed $t\in(0,1)$,
	$p_t$ is strictly positive and $C^1$. The posterior mean
	$x\mapsto\mathbb E[X_\star\mid X_t=x]$ is finite and continuously differentiable on $\mathbb{R}^d$.
	Moreover, for every $x\in\mathbb R^d$,
	\begin{equation}
		\int\nabla_xp_{X_\star\mymid X_t}(x_\star\mymid x)\,\Pdata({\rm d}x_{\star})
		=\nabla_x\int p_{X_\star\mymid X_t}(x_\star\mymid x)\,\Pdata({\rm d}x_{\star})
		=\nabla_x1=0,
		\label{eq:posterior-normalization-differentiation}
	\end{equation}
	and
	\begin{equation}
		\begin{aligned}
		\nabla_x\mathbb E[X_\star\mid X_t=x]
		&=\int x_\star
		\left[\nabla_xp_{X_\star\mymid X_t}(x_\star\mymid x)\right]^\top\Pdata({\rm d}x_{\star})=\int (x_\star-x)
		\left[\nabla_xp_{X_\star\mymid X_t}(x_\star\mymid x)\right]^\top\Pdata({\rm d}x_{\star}).
		\end{aligned}
		\label{eq:posterior-mean-differentiation}
	\end{equation}
	In particular,
	\begin{equation}\label{eq:v_t}
	v_t(x)=\frac{\mathbb E[X_\star\mid X_t=x]-x}{1-t}
	\end{equation}
	is $C^1$. The two functions
	\[
	x\longmapsto\mathbb E[\|Z-\mathbb E[Z]\|_2^2\mid X_t=x],
	\qquad
	x\longmapsto\mathbb E[\|s(Z)\|_2^2\mid X_t=x]
	\]
	are finite and continuous. Consequently, $T_n$ is $C^1$ and
	$G_n$ is closed.
\end{lemma}

\begin{lemma}[Injectivity on the restricted set]
	\label{lem:restriction}
	Under Assumptions~\ref{ass:source} and~\ref{ass:schedule}, for every
	$1\leq n\leq N$, the map $T_n$ is injective on $G_n$. Moreover,
	for every $x\in G_n$,
	\begin{equation}
		(t_n-t_{n-1})\|\nabla_xv_{t_{n-1}}(x)\|_{\mathrm{op}}\leq\frac58,
		\qquad \det\nabla_xT_n(x)>0.
		\label{eq:good-jacobian}
	\end{equation}
\end{lemma}
The proofs of Lemmas~\ref{lem:source-only-regularity} and \ref{lem:restriction} are postponed to Sections~\ref{app:posterior-regularity} and~\ref{app:restriction-proof}, respectively.

For any event $A\subseteq\mathbb R^d$, Lemmas~\ref{lem:source-only-regularity} and~\ref{lem:restriction} allow us to change variables on $G_n\cap T_n^{-1}(A)$ and obtain
\begin{equation}
	\begin{aligned}
		\int_A p_{t_n}(y)\,{\rm d}y
		&\geq \int_{A\cap T_n(G_n)} p_{t_n}(y)\,{\rm d}y
		=\int_{G_n\cap T_n^{-1}(A)}
		p_{t_n}(T_n(x))\det\nabla T_n(x)\,{\rm d}x.
	\end{aligned}
	\label{eq:change-of-variables}
\end{equation}
Since $\widehat Y_n=T_n(X_{t_{n-1}})$, we know that
\begin{align*}
	\mathbb P(\widehat Y_n\in A)-\int_Ap_{t_n}(y)\,{\rm d}y &\leq\mathbb P(X_{t_{n-1}}\notin G_n) + \mathbb P(\widehat Y_n\in A,X_{t_{n-1}} \in G_n) - \int_Ap_{t_n}(y)\,{\rm d}y\\
	&\leq\mathbb P(X_{t_{n-1}}\notin G_n)+\int_{G_n\cap T_n^{-1}(A)}
	\bigl[p_{t_{n-1}}(x)-p_{t_n}(T_n(x))
	\det\nabla T_n(x)\bigr]\,{\rm d}x\\
	&\leq\mathbb P(X_{t_{n-1}}\notin G_n)+\int_{G_n}
	\bigl[p_{t_{n-1}}(x)-p_{t_n}(T_n(x))
	\det\nabla T_n(x)\bigr]_+\,{\rm d}x.
\end{align*}
Taking the supremum over $A$ yields
\begin{align}
	\operatorname{TV}(\widehat P_n,P_{t_n})
	&\leq\mathbb P(X_{t_{n-1}}\notin G_n)+\int_{G_n}p_{t_{n-1}}(x)\left[1-\frac{p_{t_n}(T_n(x))\det\nabla T_n(x)}{p_{t_{n-1}}(x)}\right]_+\,{\rm d}x.
	\label{eq:restricted-tv}
\end{align}
In the following, we will provide a lower bound for $\frac{p_{t_n}(T_n(x))\det\nabla T_n(x)}{p_{t_{n-1}}(x)}$.
\paragraph{Step 2: Expanding the Jacobian determinant.} First, we deal with $\det\nabla_x T_n(x)$. Recall that
\begin{align}\label{eq:connection_T_u}
	\nabla_x T_n(x) = I_d + (t_n - t_{n-1})\nabla_xv_{t_{n-1}}(x).
\end{align}
In view of \eqref{eq:good-jacobian} in
Lemma~\ref{lem:restriction}, we have
$\|\nabla T_n(x)-I_d\|_{\mathrm{op}}\leq5/8$ and
$\det\nabla T_n(x)>0$ for every $x\in G_n$.
For any real matrix $E$ with $\|E\|_{\mathrm{op}}\le5/8$,
the convergent logarithmic series tells us that
\begin{align}
	|\log\det(I_d+E)-\operatorname{Tr}E|
	&\le\sum_{\ell=2}^{\infty}
	\frac{|\operatorname{Tr}(E^\ell)|}{\ell} \le \sum_{\ell=2}^{\infty}
	\frac{|\langle E^\top, E^{\ell-1}\rangle|}{\ell} \le \sum_{\ell=2}^{\infty}
	\frac{\|E\|_{\rm F}\|E^{\ell-1}\|_{\rm F}}{\ell}\notag\\
	&\le\|E\|_{\mathrm F}^2
	\sum_{\ell=2}^{\infty}\frac{\|E\|_{\mathrm{op}}^{\ell-2}}{\ell}
	\le\frac{\|E\|_{\mathrm F}^2}{2(1-\|E\|_{\mathrm{op}})}
	\le\frac43\|E\|_{\mathrm F}^2 .
	\label{eq:logdet-series}
\end{align}
For any $x \in G_n$, setting $E = \nabla T_n(x)-I_d$ in \eqref{eq:logdet-series} and using \eqref{eq:connection_T_u} yield
\begin{align}\label{eq:euler-logdet-remainder}
	\left|\log\left(\det\nabla_x T_n(x)\right) - (t_n - t_{n-1})\operatorname{Tr}(\nabla_xv_{t_{n-1}}(x))\right| \leq \frac{4}{3}(t_n - t_{n-1})^2\left\|\nabla_xv_{t_{n-1}}(x)\right\|_{\rm F}^2.
\end{align}

\paragraph{Step 3: Expanding the density ratio.}
For any fixed $n$, let $y_{n-1}$ denote a fixed
point and $y_n=T_n(y_{n-1})$, where
\begin{align*}
	T_n(x)&=x+(t_n-t_{n-1})v_{t_{n-1}}(x).
\end{align*}
Then we know the density $p_{t_n}(y)$ at $y = y_n$ is
\begin{align}
	p_{t_n}(y_n)&=\frac1{(1-t_n)^d}\int
	\psource\!\left(\frac{y_n-t_nx_\star}{1-t_n}\right)
	\Pdata({\rm d}x_{\star})\notag\\
	&=
	\left(\frac{1-t_{n-1}}{1-t_n}\right)^d
	p_{t_{n-1}}(y_{n-1})
	\int p_{X_\star\mid X_{t_{n-1}}}(x_\star\mid y_{n-1})
	\frac{\psource\bigl((y_n-t_nx_\star)/(1-t_n)\bigr)}
	{\psource\bigl((y_{n-1}-t_{n-1}x_\star)/(1-t_{n-1})\bigr)}
	\,\Pdata({\rm d}x_{\star}),
	\label{eq:density}
\end{align}
where the last equation holds due to Bayes' formula
\begin{align*}
	p_{X_\star\mid X_{t_{n-1}}}(x_\star\mid y_{n-1})=\frac{\psource\!\left(\frac{y_{n-1}-t_{n-1}x_\star}{1-t_{n-1}}\right)}{(1-t_{n-1})^d p_{t_{n-1}}(y_{n-1})}.
\end{align*}
We collect the identities and bounds for the velocity Jacobian used below.
\begin{lemma}[Velocity Jacobian]\label{lem:velocity-jacobian}
Under Assumption~\ref{ass:source}, for every $t\in(0,1)$ and $x\in\mathbb R^d$,
conditional on $X_t=x$, one has
\begin{align}
	t(X_\star-\mathbb E[X_\star\mid X_t=x])
	&=-(1-t)(Z-\mathbb E[Z\mid X_t=x]).
	\label{eq:centered-interpolant-identity}
\end{align}
Moreover,
\begin{align}
	\nabla_xv_t(x)
	&=-\frac{1}{1-t}I-\frac{1}{t(1-t)}
	\mathbb E\!\left[(Z-\mathbb E[Z\mid X_t=x])s(Z)^\top
	\,\middle|\,X_t=x\right].
	\label{eq:gradient_v_t_another_form}
\end{align}
In addition,
\begin{align}
	\left\|\nabla v_t(x)+\frac{I_d}{1-t}\right\|_{\mathrm F}^2
	&\le\frac{
		\mathbb E[\|Z-\mathbb E[Z]\|_2^2\mid X_t=x]\,
		\mathbb E[\|s(Z)\|_2^2\mid X_t=x]}
	{t^2(1-t)^2},\label{eq:jacobian-cs}\\
	\|\nabla v_t(x)\|_{\mathrm F}^2
	&\le\frac{2d}{(1-t)^2}
	+\frac{2\mathbb E[\|Z-\mathbb E[Z]\|_2^2\mid X_t=x]\,
		\mathbb E[\|s(Z)\|_2^2\mid X_t=x]}
	{t^2(1-t)^2}.
	\label{eq:jacobian-frobenius}
\end{align}
\end{lemma}
The proof is given in Appendix~\ref{app:velocity-jacobian}.

For the integration variable
$x_\star$, we let $z=(y_{n-1}-t_{n-1}x_\star)/(1-t_{n-1})$.
Applying \eqref{eq:centered-interpolant-identity} at $t=t_{n-1}$ and $x=y_{n-1}$ gives
\begin{align}
	&\frac{y_n-t_nx_\star}{1-t_n}
	-\frac{y_{n-1}-t_{n-1}x_\star}{1-t_{n-1}}
	\notag\\
	&\quad\stackrel{\eqref{eq:v_t}}{=} -\frac{t_n-t_{n-1}}{(1-t_n)(1-t_{n-1})}x_\star + \frac{y_{n-1}}{1-t_n} + \frac{t_n-t_{n-1}}{1-t_n}\mathbb{E}\left[\frac{X_\star - y_{n-1}}{1 - t_{n-1}}~\Big|~X_{t_{n-1}} = y_{n-1}\right]\notag\\
	&\qquad -\frac{y_{n-1}}{1-t_{n-1}}\notag\\
	&\quad=-\frac{t_n-t_{n-1}}{(1-t_n)(1-t_{n-1})}
	\left(x_\star-\mathbb E[X_\star\mid X_{t_{n-1}}=y_{n-1}]\right)
	\notag\\
	&\quad=\frac{t_n-t_{n-1}}{t_{n-1}(1-t_n)}
	\left(z-\mathbb E[Z\mid X_{t_{n-1}}=y_{n-1}]\right).
	\label{eq:displacement}
\end{align}
Since $s(\cdot)$ is $L$-Lipschitz, by Taylor's expansion, we have
\begin{align}
	&\log \psource\left(\frac{y_n-t_nx_\star}{1-t_n}\right)
	-\log \psource\left(\frac{y_{n-1}-t_{n-1}x_\star}{1-t_{n-1}}\right)
	\notag\\
	&\qquad=
	-\frac{t_n-t_{n-1}}{(1-t_n)(1-t_{n-1})}
	s(z)^\top
	\left(x_\star-\mathbb E[X_\star\mid X_{t_{n-1}}=y_{n-1}]\right)
	+r_n^{(2)}(x_\star),
	\label{eq:source-expansion}
\end{align}
where the remainder term satisfies
\begin{align}
	|r_n^{(2)}(x_\star)|
	&\le
	\frac{L(t_n-t_{n-1})^2}{2t_{n-1}^2(1-t_n)^2}
	\left\|z-\mathbb E[Z\mid X_{t_{n-1}}=y_{n-1}]\right\|_2^2 .
	\label{eq:source-remainder}
\end{align}
We define
\begin{align}
	\xi_n
	&:=
	s(Z)^\top\left(X_\star-\mathbb E[X_\star\mid X_{t_{n-1}}=y_{n-1}]\right)
	-
	\mathbb E\left[
	s(Z)^\top\left(X_\star-\mathbb E[X_\star\mid X_{t_{n-1}}=y_{n-1}]\right)
	\,\middle|\,X_{t_{n-1}}=y_{n-1}\right].
	\label{eq:xi}
\end{align}
Then
\begin{align}\label{eq:zero_mean_xi}
	\mathbb{E}[\xi_n\,|\,X_{t_{n-1}}=y_{n-1}] = 0.
\end{align}
Combining \eqref{eq:density}, \eqref{eq:centered-interpolant-identity}, \eqref{eq:source-expansion} and \eqref{eq:xi}, one arrives at
\begin{align}
	&\log\frac{p_{t_n}(y_n)}{p_{t_{n-1}}(y_{n-1})}\notag\\
 &\quad= d\log\left(\frac{1-t_{n-1}}{1-t_n}\right) - \frac{t_n-t_{n-1}}{(1-t_n)(1-t_{n-1})}\mathbb E\left[
	s(Z)^\top\left(X_\star-\mathbb E[X_\star\mid X_{t_{n-1}}=y_{n-1}]\right)
	\,\middle|\,X_{t_{n-1}}=y_{n-1}\right]\notag\\
	&\qquad + \log\mathbb E\left[\exp\left(- \frac{t_n-t_{n-1}}{(1-t_n)(1-t_{n-1})}\xi_n + r_n^{(2)}(X_{\star})\right)
	\,\middle|\,X_{t_{n-1}}=y_{n-1}\right]\notag\\
	&\quad=d\log\left(\frac{1-t_{n-1}}{1-t_n}\right) + \frac{t_n-t_{n-1}}{(1-t_n)t_{n-1}}\mathbb E\left[
	s(Z)^\top\left(Z-\mathbb E[Z\mid X_{t_{n-1}}=y_{n-1}]\right)
	\,\middle|\,X_{t_{n-1}}=y_{n-1}\right]\notag\\
	&\qquad + \log\mathbb E\left[\exp\left(- \frac{t_n-t_{n-1}}{(1-t_n)(1-t_{n-1})}\xi_n + r_n^{(2)}(X_{\star})\right)
	\,\middle|\,X_{t_{n-1}}=y_{n-1}\right].\label{eq:density_ratio}
\end{align}

\paragraph{Step 4: Controlling $\log\frac{p_{t_n}(T_n(x))\det\nabla T_n(x)}{p_{t_{n-1}}(x)}$.} 
Taking the trace in \eqref{eq:gradient_v_t_another_form} of Lemma~\ref{lem:velocity-jacobian} gives
\begin{align*}
	 &\operatorname{Tr}(\nabla v_{t_{n-1}}(y_{n-1}))= -\frac{d}{1-t_{n-1}} - \frac{1}{t_{n-1}(1-t_{n-1})}\mathbb{E}\left[s\left(Z\right)^\top(Z-\mathbb E[Z\mid X_{t_{n-1}}=y_{n-1}])~\big|~X_{t_{n-1}} = y_{n-1}\right].
\end{align*}
Combining \eqref{eq:euler-logdet-remainder}, \eqref{eq:density_ratio}, the previous equation and the fact
$\frac{t_n-t_{n-1}}{t_{n-1}}\left(\frac{1}{1-t_n}-\frac{1}{1-t_{n-1}}\right)
    =\frac{(t_n-t_{n-1})^2}{t_{n-1}(1-t_{n-1})(1-t_n)}$,
we have, for any $y_{n-1}\in G_n$,
\begin{align}
	&\log\left(\frac{p_{t_n}(y_n)\det\nabla T_n(y_{n-1})}{p_{t_{n-1}}(y_{n-1})}\right)\notag\\
	&\quad\geq -\frac{d(t_n-t_{n-1})}{1-t_{n-1}} + d\log\left(\frac{1-t_{n-1}}{1-t_n}\right)\notag\\
 &\qquad+ \frac{(t_n - t_{n-1})^2}{t_{n-1}(1-t_{n-1})(1-t_n)}\mathbb{E}\left[s\left(Z\right)^\top(Z-\mathbb E[Z\mid X_{t_{n-1}}=y_{n-1}])~\big|~X_{t_{n-1}} = y_{n-1}\right]\notag\\
	&\qquad - \frac{4}{3}(t_n - t_{n-1})^2\left\|\nabla v_{t_{n-1}}(y_{n-1})\right\|_{\rm F}^2\notag\\
 &\qquad+ \log\mathbb E\left[\exp\left(- \frac{t_n-t_{n-1}}{(1-t_n)(1-t_{n-1})}\xi_n + r_n^{(2)}(X_{\star})\right)
	\,\middle|\,X_{t_{n-1}}=y_{n-1}\right].\label{ineq:lower_log_quantity}
\end{align}
Since $s(\cdot)$ is $L$-Lipschitz, the following lower bound holds:
\begin{align}
	&\mathbb{E}\left[s\left(Z\right)^\top(Z-\mathbb E[Z\mid X_{t_{n-1}}=y_{n-1}])~\big|~X_{t_{n-1}} = y_{n-1}\right]\notag\\
	&\quad = \mathbb{E}\left[\big(s(Z)-s(\mathbb E[Z\mid X_{t_{n-1}}=y_{n-1}])\big)^\top(Z-\mathbb E[Z\mid X_{t_{n-1}}=y_{n-1}])~\big|~X_{t_{n-1}} = y_{n-1}\right]\notag\\
	&\quad \geq -L\mathbb{E}\left[\|Z-\mathbb E[Z\mid X_{t_{n-1}}=y_{n-1}]\|_2^2~\big|~X_{t_{n-1}} = y_{n-1}\right].\label{ineq:lower_bound_product}
\end{align}
Jensen's inequality also tells us
\begin{align}
	&\log\mathbb E\left[\exp\left(- \frac{t_n-t_{n-1}}{(1-t_n)(1-t_{n-1})}\xi_n + r_n^{(2)}(X_{\star})\right)
	\,\middle|\,X_{t_{n-1}}=y_{n-1}\right]\notag\\
	&\quad\geq \mathbb E\left[- \frac{t_n-t_{n-1}}{(1-t_n)(1-t_{n-1})}\xi_n + r_n^{(2)}(X_{\star})
	\,\middle|\,X_{t_{n-1}}=y_{n-1}\right]\notag\\
	&\quad\stackrel{\eqref{eq:zero_mean_xi}}{=} \mathbb E\left[r_n^{(2)}(X_{\star})
	\,\middle|\,X_{t_{n-1}}=y_{n-1}\right]\notag\\
	&\quad \stackrel{\eqref{eq:source-remainder}}{\geq} -\frac{L(t_n-t_{n-1})^2}{2t_{n-1}^2(1-t_n)^2}
	\mathbb E\left[\left\|Z-\mathbb E[Z\mid X_{t_{n-1}}=y_{n-1}]\right\|_2^2
	\,\middle|\,X_{t_{n-1}}=y_{n-1}\right].\label{ineq:Jensen_log_inequality}
\end{align}
Putting \eqref{ineq:lower_log_quantity}, \eqref{ineq:lower_bound_product} and \eqref{ineq:Jensen_log_inequality} together, we obtain that for any $x \in G_n$,
\begin{align}
	&\log\left(\frac{p_{t_n}(T_n(x))\det\nabla T_n(x)}{p_{t_{n-1}}(x)}\right)\notag\\
	&\quad \geq -\frac{3L(t_n-t_{n-1})^2}{2t_{n-1}^2(1-t_n)^2}
	\mathbb{E}\left[\|Z-\mathbb E[Z\mid X_{t_{n-1}}=x]\|_2^2~\big|~X_{t_{n-1}} = x\right]\notag\\
 &\qquad- \frac{4}{3}(t_n - t_{n-1})^2\left\|\nabla v_{t_{n-1}}(x)\right\|_{\rm F}^2\notag\\
	&\quad \geq -3\frac{(t_n - t_{n-1})^2}{t_{n-1}^2(1-t_n)^2}\biggl(d + L\mathbb{E}\left[\|Z-\mathbb E[Z]\|_2^2~\big|~X_{t_{n-1}} = x\right]\notag\\
 &\hspace{4em}+ \mathbb E[\|Z-\mathbb E[Z]\|_2^2\mid X_{t_{n-1}}=x]\,
	\mathbb E[\|s(Z)\|_2^2\mid X_{t_{n-1}}=x]\biggr).\label{eq:ratio-bound}
\end{align}
Here, the first inequality comes from $-a - \log(1-a) \geq 0$ with $a=(t_n-t_{n-1})/(1-t_{n-1})$ and $t_{n-1}(1-t_n)/(1-t_{n-1}) \leq 1$ by Assumption~\ref{ass:schedule}.
The second inequality uses \eqref{eq:jacobian-frobenius} and the fact that $\mathbb E[Z\mid X_{t_{n-1}}=x]$ minimizes the conditional mean squared error.

Finally, inserting \eqref{eq:ratio-bound} into
\eqref{eq:restricted-tv} and using $[1-e^{-v}]_+\le v$ for $v \ge 0$ yields
\begin{align}
	\operatorname{TV}(P_{\widehat Y_n},P_{t_n})
	&\le\mathbb P(X_{t_{n-1}}\notin G_n)
	+3\frac{(t_n-t_{n-1})^2}{t_{n-1}^2(1-t_n)^2}
	\Bigl\{d+L\mathbb E\bigl[\|Z-\mathbb E[Z]\|_2^2\bigr]\notag\\
	&\qquad+
	\mathbb E\Bigl[
	\mathbf 1_{\{X_{t_{n-1}}\in G_n\}}
	\mathbb E[\|Z-\mathbb E[Z]\|_2^2\mid X_{t_{n-1}}]\,
	\mathbb E[\|s(Z)\|_2^2\mid X_{t_{n-1}}]
	\Bigr]\Bigr\}.
	\label{eq:one-step}
\end{align}

\paragraph{Step 5: Putting it all together.}
 By virtue of \eqref{eq:source-score-norm-mgf} in
Lemma~\ref{lem:source-estimates}, conditional Jensen's
inequality and the tower property, one has
\begin{align*}
	&\mathbb E\left[\exp\left\{
	\frac{\mathbb E[\|s(Z)\|_2^2\mid X_{t_{n-1}}]}{4L}\right\}\right] \leq
	\mathbb E\left[\mathbb E\left[
	\exp\left\{\frac{\|s(Z)\|_2^2}{4L}\right\}
	\,\middle|\,X_{t_{n-1}}\right]\right]=\mathbb E\left[\exp\left\{\frac{\|s(Z)\|_2^2}{4L}\right\}\right]
	\leq2^{d/2}.
\end{align*}
Applying Markov's inequality to this exponential yields
\begin{align}
	&\mathbb P\!\left(
	\mathbb E[\|s(Z)\|_2^2\mid X_{t_{n-1}}]
	>4L\{d+12\log(N+1)\}\right)\notag\\
	&\qquad=\mathbb P\!\left(
	\exp\left\{\frac{\mathbb E[\|s(Z)\|_2^2\mid X_{t_{n-1}}]}{4L}\right\}
	>e^{d+12\log(N+1)}\right)\notag\\
	&\qquad\leq e^{-d-12\log(N+1)}
	\mathbb E\left[\exp\left\{
	\frac{\mathbb E[\|s(Z)\|_2^2\mid X_{t_{n-1}}]}{4L}\right\}\right]\notag\\
	&\qquad\leq
	\exp\!\left(\frac d2\log2-d-12\log(N+1)\right)\notag\\
	&\qquad\leq(N+1)^{-12}.
	\label{eq:score-tail}
\end{align}
In addition, Markov's inequality implies that
\begin{align}
	&\mathbb P\!\left(
	\mathbb E[\|Z-\mathbb E[Z]\|_2^2\mid X_{t_{n-1}}]
	>\frac{t_{n-1}^2(1-t_n)^2}
	{256L(t_n-t_{n-1})^2\{d+12\log(N+1)\}}\right)\notag\\
	&\qquad\leq
	\frac{256L(t_n-t_{n-1})^2\{d+12\log(N+1)\}}
	{t_{n-1}^2(1-t_n)^2}
	\mathbb E\left[\mathbb E[\|Z-\mathbb E[Z]\|_2^2
	\mid X_{t_{n-1}}]\right]\notag\\
	&\qquad=
	\frac{256L(t_n-t_{n-1})^2\{d+12\log(N+1)\}}
	{t_{n-1}^2(1-t_n)^2}\mathbb E\left[\|Z-\mathbb E[Z]\|_2^2\right].\label{eq:conditional-exp-tail}
\end{align}
In view of \eqref{eq:good-set}, \eqref{eq:score-tail}, \eqref{eq:conditional-exp-tail} and the union bound, we have
\begin{align}
	\mathbb P(X_{t_{n-1}}\notin G_n)
	&\leq(N+1)^{-12}+
	\frac{256L(t_n-t_{n-1})^2\{d+12\log(N+1)\}}
	{t_{n-1}^2(1-t_n)^2}\mathbb E\left[\|Z-\mathbb E[Z]\|_2^2\right].
	\label{eq:second-discarded}
\end{align}

On the event
$\{X_{t_{n-1}}\in G_n\}$, its conditional score factor $\mathbb E[\|s(Z)\|_2^2\mid X_{t_{n-1}}=x]$ is at most
$4L\{d+12\log(N+1)\}$, and consequently
\begin{align}
	&\mathbb E\!\left[\mathbf1_{G_n}(X_{t_{n-1}})
	\mathbb E[\|Z-\mathbb E[Z]\|_2^2\mid X_{t_{n-1}}]
	\mathbb E[\|s(Z)\|_2^2\mid X_{t_{n-1}}]\right]\notag\\
	&\qquad\leq4L\{d+12\log(N+1)\}
	\mathbb E\!\left[\mathbf1_{G_n}(X_{t_{n-1}})
	\mathbb E[\|Z-\mathbb E[Z]\|_2^2\mid X_{t_{n-1}}]\right]\notag\\
	&\qquad\leq4L\{d+12\log(N+1)\}
	\mathbb E\!\left[
	\mathbb E[\|Z-\mathbb E[Z]\|_2^2\mid X_{t_{n-1}}]\right]\notag\\
	&\qquad=4L\{d+12\log(N+1)\}
	\mathbb E\left[\|Z-\mathbb E[Z]\|_2^2\right].
	\label{eq:second-product}
\end{align}
Substituting \eqref{eq:second-discarded} and
\eqref{eq:second-product} into \eqref{eq:one-step} gives
\begin{align*}
	\mathsf{TV}(P_{\widehat Y_n},P_{t_n})
	&\leq(N+1)^{-12}+\frac{256L(t_n-t_{n-1})^2\{d+12\log(N+1)\}}
	{t_{n-1}^2(1-t_n)^2}\mathbb E\left[\|Z-\mathbb E[Z]\|_2^2\right]\\
	&\quad+\frac{3(t_n-t_{n-1})^2}{t_{n-1}^2(1-t_n)^2}
	\Bigl\{d+L\mathbb E\left[\|Z-\mathbb E[Z]\|_2^2\right]+4L\{d+12\log(N+1)\}
	\mathbb E\left[\|Z-\mathbb E[Z]\|_2^2\right]\Bigr\}\\
	&\leq (N+1)^{-12}
	+\frac{(t_n-t_{n-1})^2}{t_{n-1}^2(1-t_n)^2}
	\Bigl\{4d+4L\mathbb E\left[\|Z-\mathbb E[Z]\|_2^2\right]+300L\{d+12\log(N+1)\}
	\mathbb E\left[\|Z-\mathbb E[Z]\|_2^2\right]\Bigr\}.
\end{align*}
We know from \eqref{eq:scale-lower-bounds} in
Lemma~\ref{lem:source-estimates} that
$d\leq L\mathbb E\bigl[\|Z-\mathbb E[Z]\|_2^2\bigr]$. Therefore, we have
\begin{align*}
	4d+4L\mathbb E\bigl[\|Z-\mathbb E[Z]\|_2^2\bigr]
	&\leq8L\mathbb E\bigl[\|Z-\mathbb E[Z]\|_2^2\bigr]\leq8L\{d+12\log(N+1)\}\mathbb E\bigl[\|Z-\mathbb E[Z]\|_2^2\bigr].
\end{align*}
Combining these inequalities gives
\begin{align}
	\mathsf{TV}(P_{\widehat Y_n},P_{t_n})
	&\leq(N+1)^{-12}+\frac{CL(t_n-t_{n-1})^2\{d+\log(N+1)\}}
	{t_{n-1}^2(1-t_n)^2}\mathbb E\left[\|Z-\mathbb E[Z]\|_2^2\right],\label{ineq:one-step}
\end{align}
The grid estimate \eqref{eq:time-grid-sums} with $r=2$ gives
\begin{align*}
	\sum_{n=1}^N\frac{(t_n-t_{n-1})^2}{t_{n-1}^2(1-t_n)^2}
	\leq\frac{c_s^2\log^2(N+1)}{N+1}.
\end{align*}
Summing the one-step bound \eqref{ineq:one-step},
we obtain
\begin{align*}
	\sum_{n=1}^N\mathsf{TV}(P_{\widehat Y_n},P_{t_n})
	&\leq\frac{1}{(N+1)^{11}}+CL\{d+\log(N+1)\}\frac{\log^2(N+1)}{N+1}\mathbb E\left[\|Z-\mathbb E[Z]\|_2^2\right].
\end{align*}
Putting
\eqref{eq:TV-telescope}, 
\eqref{eq:init-absorbed} and the previous inequality together, we arrive at
\begin{align*}
	\mathsf{TV}(P_{Y_N},P_{t_N})
	&\leq\frac{\sqrt{L\mathbb E\left[\|Z-\mathbb E[Z]\|_2^2\right]}}{(N+1)^2}
	+(N+1)^{-11}+CL\{d+\log(N+1)\}\frac{\log^2(N+1)}{N+1}
	\mathbb E\left[\|Z-\mathbb E[Z]\|_2^2\right].
\end{align*}
In addition, \eqref{eq:scale-lower-bounds} tells us
$$
1\leq \sqrt{L\mathbb E\left[\|Z-\mathbb E[Z]\|_2^2\right]}
\leq L\mathbb E\left[\|Z-\mathbb E[Z]\|_2^2\right].
$$
As a consequence, one has
\begin{align*}
	&\frac{\sqrt{L\mathbb E\left[\|Z-\mathbb E[Z]\|_2^2\right]}}{(N+1)^2}
	+(N+1)^{-11}\leq\frac{2L\mathbb E\left[\|Z-\mathbb E[Z]\|_2^2\right]}{(N+1)^2}.
\end{align*}
Therefore, we have
\begin{align*}
	\mathsf{TV}(P_{Y_N},P_{t_N})
	&\lesssim L(d+\log N)\mathbb E\left[\|Z-\mathbb E[Z]\|_2^2\right]
	\frac{\log^2N}{N},
\end{align*}
which finishes the proof of Theorem~\ref{thm:general-moments}.

\section{Discussion}
\label{sec:discussion}

In this work, we have established the first non-asymptotic convergence guarantee for the Euler sampler in flow matching with a general non-Gaussian source. Compared with prior work, our analysis does not require boundedness assumptions on the velocity field or additional tail conditions on the target distribution. We assume only that the source distribution has a continuously differentiable density that is positive everywhere on $\mathbb{R}^d$, a finite second moment, and a Lipschitz score function. These mild assumptions accommodate, among many other examples, correlated Gaussian distributions, multivariate Student's $t$ distributions, and Gaussian mixtures with a shared covariance. To establish our theory, we have developed new proof techniques for handling nonlinear source score functions and proving restricted injectivity of the Euler map, which extend beyond the global injectivity available for Gaussian sources. These techniques may be of independent interest.

The theory of flow matching with general source distributions remains in its early stages, leaving a number of important questions open.  For instance, our current analysis assumes access to the exact population velocity field; extending it to account for velocity estimation error and quantifying its effect on sampling accuracy are natural next steps.  It is also unclear whether particular source choices, such as multivariate Student's \(t\) distributions, can accelerate convergence for heavy-tailed targets. Another related question concerns the choice of source-target coupling. In image-to-image and text-to-image generation, many approaches train on paired source and target samples to preserve the desired correspondence during sampling. Developing convergence guarantees for dependent source-target couplings could clarify which couplings lead to faster sampling.  Another direction is to extend our theory from continuous distributions on Euclidean space to discrete and Riemannian flow matching, where generating structured data is a central task.


\section*{Acknowledgments}
Y.~Chen is supported in part by the Alfred P.~Sloan Research Fellowship,  the ONR grant N00014-25-1-2344, the AFOSR grant FA9550261B178, and 
the Wharton AI \& Analytics Initiative's AI Research Fund. 
G.~Li is supported in part by the Chinese University of Hong Kong Direct Grant for Research and the Hong Kong Research Grants Council ECS 2191363. This work is also supported in part by the NSF under Cooperative Agreement No. 2433450.

\appendix

\section{Proofs of auxiliary lemmas}

\subsection{Proof of examples in Section~\ref{eq:theory-implications}}
\label{app:source-examples}

Note that all three examples of source distributions have strictly positive $C^1$ densities on $\mathbb R^d$. We develop upper bounds for $L\Vsource$ below. For each source, we shall upper bound the Lipschitz constant $L$ by $\sup_x\|\nabla s(x)\|_{\mathrm{op}}$. To see this, we observe that

\begin{align*}
\|s(z)-s(z')\|_2
&=\left\|\int_0^1\nabla s(z'+\theta(z-z'))(z-z')\,{\rm d}\theta\right\|_2\\
&\leq\int_0^1\|\nabla s(z'+\theta(z-z'))\|_{\mathrm{op}}\|z-z'\|_2\,{\rm d}\theta
\leq \left(\sup_x\|\nabla s(x)\|_{\mathrm{op}}\right)\|z-z'\|_2.
\end{align*}

\paragraph{Gaussian source.}
For $Z\sim\mathcal N(\mu,\Sigma)$ with $\Sigma\succ0$, one can show that
\begin{align*}
s(z)&=\nabla_z\left[-\frac12(z-\mu)^\top\Sigma^{-1}(z-\mu)\right]
=-\Sigma^{-1}(z-\mu),\\
\nabla s(z)&=-\Sigma^{-1},\qquad
L=\|\Sigma^{-1}\|_{\mathrm{op}}=\frac1{\lambda_{\min}(\Sigma)},\\
\Vsource &=\operatorname{tr}\mathbb E[(Z-\mu)(Z-\mu)^\top]
=\operatorname{tr}(\Sigma).
\end{align*}
Therefore, we have $L\Vsource=\frac{\operatorname{tr}(\Sigma)}{\lambda_{\min}(\Sigma)}$.

\paragraph{Multivariate Student-$t$ source.}
Let $\nu>2$ and let $\Sigma\succ0$ be the scale matrix. Set
$y=\Sigma^{-1/2}(z-\mu)$ and $q=\|y\|_2^2$. It is straightforward to show the following:
\begin{align*}
\psource(z)&\propto |\Sigma|^{-1/2}(1+q/\nu)^{-(\nu+d)/2},
\qquad \nabla_zq=2\Sigma^{-1}(z-\mu),\\
s(z)&=-\frac{\nu+d}{2}\frac{\nabla_zq}{\nu+q}
=-\frac{\nu+d}{\nu+q}\Sigma^{-1}(z-\mu),\\
\nabla s(z)&=-\frac{\nu+d}{\nu+q}\Sigma^{-1}
+\frac{2(\nu+d)}{(\nu+q)^2}\Sigma^{-1}(z-\mu)(z-\mu)^\top\Sigma^{-1}=-\frac{\nu+d}{\nu+q}\Sigma^{-1/2}
\left(I_d-\frac{2yy^\top}{\nu+q}\right)\Sigma^{-1/2}.
\end{align*}
Since
\[
0\preceq yy^\top\preceq qI_d,\qquad
-I_d\preceq\frac{\nu-q}{\nu+q}I_d
\preceq I_d-\frac{2yy^\top}{\nu+q}\preceq I_d,
\]
we obtain the Lipschitz constant as follows
\begin{align*}
\|\nabla s(z)\|_{\mathrm{op}}
&\leq\frac{\nu+d}{\nu+q}\|\Sigma^{-1/2}\|_{\mathrm{op}}^2
\left\|I_d-\frac{2yy^\top}{\nu+q}\right\|_{\mathrm{op}}\leq\frac{\nu+d}{(\nu+q)\lambda_{\min}(\Sigma)}
\leq\frac{\nu+d}{\nu\lambda_{\min}(\Sigma)}=:L.
\end{align*}
Using independent $G\sim\mathcal N(0,I_d)$ and $S\sim\chi^2_\nu$, one can write
\begin{align*}
Z&\overset{\mathrm d}=\mu+\sqrt{\nu/S}\,\Sigma^{1/2}G,\qquad
\mathbb E[Z]=\mathbb E\left[\mu+\sqrt{\nu/S}\,\Sigma^{1/2}\mathbb E[G\mid S]\right]=\mu.
\end{align*}
Hence, we can bound the total variance of $Z$ as follows
\begin{align*}
\Vsource &=\nu\mathbb E[S^{-1}G^\top\Sigma G]
=\nu\mathbb E[S^{-1}]\operatorname{tr}(\Sigma\mathbb E[GG^\top])
=\frac{\nu}{\nu-2}\operatorname{tr}(\Sigma).
\end{align*}
Therefore, we have
\begin{align*}
L\Vsource &=\frac{\nu+d}{\nu\lambda_{\min}(\Sigma)}
\frac{\nu\operatorname{tr}(\Sigma)}{\nu-2}
=\frac{\nu+d}{\nu-2}\frac{\operatorname{tr}(\Sigma)}{\lambda_{\min}(\Sigma)}.
\end{align*}

\paragraph{Gaussian mixture with a common covariance.}
Let $\varphi_\Sigma$ be the $\mathcal N(0,\Sigma)$ density and define
\[
w_k(z)=\frac{\pi_k\varphi_\Sigma(z-\mu_k)}{\psource(z)}=\frac{\pi_k\varphi_\Sigma(z-\mu_k)}{\sum_{l=1}^m \pi_l\varphi_\Sigma(z-\mu_l)},\qquad
m_z=\sum_k w_k(z)\mu_k.
\]
We can then calculate the score function as follows:
\begin{align*}
s(z)&=\frac{\sum_k\pi_k\nabla_z\varphi_\Sigma(z-\mu_k)}{\psource(z)}
=-\sum_k w_k\Sigma^{-1}(z-\mu_k)=\Sigma^{-1}(m_z-z).
\end{align*}
To bound its gradient, we note that
\begin{align*}
	\nabla w_k(z)&=w_k[-\Sigma^{-1}(z-\mu_k)-s(z)]=w_k\Sigma^{-1}(\mu_k-m_z),\\
	\nabla m_z&=\sum_k\mu_k(\nabla w_k)^\top
	=\sum_k w_k\mu_k(\mu_k-m_z)^\top\Sigma^{-1}=\sum_k w_k(\mu_k-m_z)(\mu_k-m_z)^\top\Sigma^{-1},
\end{align*}
which implies that
$$
\nabla s(z)=-\Sigma^{-1}
+\Sigma^{-1}\left[\sum_k w_k(\mu_k-m_z)(\mu_k-m_z)^\top\right]\Sigma^{-1}=\Sigma^{-1/2}(C_z-I_d)\Sigma^{-1/2},
$$
where
\[
C_z=\Sigma^{-1/2}\left[\sum_k w_k
\left(\mu_k-\sum_j w_j\mu_j\right)
\left(\mu_k-\sum_j w_j\mu_j\right)^\top\right]\Sigma^{-1/2}.
\]
For $v\in\mathbb R^d$, set
$r_k=v^\top\Sigma^{-1/2}\mu_k$, $\overline r=\sum_k w_kr_k$,
$r_-=\min_k r_k$, $r_+=\max_k r_k$, and $c=(r_-+r_+)/2$. Then
\begin{align*}
0\leq v^\top C_zv
&=\sum_k w_k(r_k-\overline r)^2
=\sum_k w_k(r_k-c)^2-(\overline r-c)^2\\ &\leq\sum_k w_k\frac{(r_+-r_-)^2}{4}
=\frac14\max_{i,j}|v^\top\Sigma^{-1/2}(\mu_i-\mu_j)|^2.
\end{align*}
By the definition of maximum component separation $\Delta$, we further have
\begin{align*}
	v^\top C_zv
&\leq\frac{\|v\|_2^2}{4}\max_{i,j}\|\Sigma^{-1/2}(\mu_i-\mu_j)\|_2^2
=\frac{\Delta^2}{4}\|v\|_2^2.
\end{align*}
Consequently,
\begin{align*}
0\preceq C_z\preceq\frac{\Delta^2}{4}I_d
&\quad\Longrightarrow\quad
-I_d\preceq C_z-I_d\preceq\left(\frac{\Delta^2}{4}-1\right)I_d,\\
\|\nabla s(z)\|_{\mathrm{op}}
&\leq\|\Sigma^{-1/2}\|_{\mathrm{op}}^2\|C_z-I_d\|_{\mathrm{op}}
\leq\frac{\max\{1,\Delta^2/4-1\}}{\lambda_{\min}(\Sigma)}=:L.
\end{align*}
Finally, to compute the total variance $\Vsource$, let $\mathbb P(K=k)=\pi_k$ and $G\sim\mathcal N(0,I_d)$ be independent. Then we have
\begin{align*}
Z&\overset{\mathrm d}=\mu_K+\Sigma^{1/2}G,\qquad
\mathbb E[Z]=\sum_k\pi_k\mu_k+\Sigma^{1/2}\mathbb E[G]=\overline\mu.
\end{align*}
Therefore,  we have
\begin{align*}
\Vsource &=\sum_k\pi_k\mathbb E\bigl[\|\mu_k-\overline\mu+\Sigma^{1/2}G\|_2^2\bigr]\\
&=\sum_k\pi_k\left[\|\mu_k-\overline\mu\|_2^2
+2(\mu_k-\overline\mu)^\top\Sigma^{1/2}\mathbb E[G]
+\operatorname{tr}(\Sigma\mathbb E[GG^\top])\right]\\
&=B+\operatorname{tr}(\Sigma),\\
L\Vsource &=\frac{\operatorname{tr}(\Sigma)+B}{\lambda_{\min}(\Sigma)}
\max\left\{1,\frac{\Delta^2}{4}-1\right\}.
\end{align*}

\subsection{Proof of Lemma~\ref{lem:grid-estimates}}\label{sec:proof_grid}
	Since $t_n>t_{n-1}$, Assumption~\ref{ass:schedule} yields
	\[
	t_n-t_{n-1}
	\leq\frac{c_s\log(N+1)}{N+1}t_{n-1}(1-t_n)
	\leq\frac{c_s\log(N+1)}{N+1}t_{n-1}(1-t_{n-1}).
	\]
	Consequently, for every $r\geq1$, 
	\begin{align*}
		\sum_{n=1}^N\left\{\frac{t_n-t_{n-1}}{t_{n-1}(1-t_n)}\right\}^r
		&\leq N\left\{\frac{c_s\log(N+1)}{N+1}\right\}^r \leq c_s^r\frac{\log^r(N+1)}{(N+1)^{r-1}}.
	\end{align*}
	This proves \eqref{eq:time-grid-sums}.

\subsection{Proof of Lemma~\ref{lem:initialization-absorbed}}\label{app:initialization}
Since
\begin{align*}
	p_{Y_0}(y)
	&=\frac1{(1-t_0)^d}\psource\!\left(\frac{y}{1-t_0}\right),~\qquad~
	p_{X_{t_0}\mid X_\star}(y\mid x_\star)
	=\frac1{(1-t_0)^d}
	\psource\!\left(\frac{y-t_0x_\star}{1-t_0}\right),
\end{align*}
we have
\begin{align*}
	\frac{p_{X_{t_0}\mid X_\star}(y\mid x_\star)}{p_{Y_0}(y)} = \frac{\frac1{(1-t_0)^d}
		\psource\!\left(\frac{y-t_0x_\star}{1-t_0}\right)}{\frac1{(1-t_0)^d}\psource\!\left(\frac{y}{1-t_0}\right)}
	=\frac{
		\psource\!\left(\frac{y-t_0x_\star}{1-t_0}\right)}{\psource\!\left(\frac{y}{1-t_0}\right)}.
\end{align*}
For fixed $z,x_\star\in\mathbb R^d$, \eqref{ineq:remainder_term} implies that
\begin{subequations}
	\begin{align}
		\log\frac{\psource(z)}{\psource\bigl(z+t_0x_\star/(1-t_0)\bigr)}
		&\leq-\frac{t_0}{1-t_0}x_\star^\top s(z)
		+\frac{Lt_0^2}{2(1-t_0)^2}\|x_\star\|_2^2,\label{eq:translated-log-density-upper}\\
		\left|\log\frac{\psource(z)}{\psource\bigl(z+t_0x_\star/(1-t_0)\bigr)}\right|
		&\leq\frac{t_0}{1-t_0}\|x_\star\|_2\|s(z)\|_2
		+\frac{Lt_0^2}{2(1-t_0)^2}\|x_\star\|_2^2.\label{eq:translated-log-density-absolute}
	\end{align}
\end{subequations}
Therefore, we can bound the KL divergence as follows:
\begin{align}
	\mathsf{KL}\bigl(P_{X_{t_0}\mid X_\star}(\cdot\mid x_\star)
	\|P_{Y_0}\bigr)
	&=\int\frac1{(1-t_0)^d}
	\psource\!\left(\frac{y-t_0x_\star}{1-t_0}\right)
	\log\frac{\psource\bigl((y-t_0x_\star)/(1-t_0)\bigr)}
	{\psource\bigl(y/(1-t_0)\bigr)}\,{\rm d}y\notag\\
	&=\int \psource(z)
	\log\frac{\psource(z)}{\psource\bigl(z+t_0x_\star/(1-t_0)\bigr)}\,{\rm d}z\notag\\
	&\leq-\frac{t_0}{1-t_0}x_\star^\top\mathbb E[s(Z)]
	+\frac{Lt_0^2}{2(1-t_0)^2}\|x_\star\|_2^2\notag\\
	&\stackrel{\eqref{eq:source-score-mgf}}{=}\frac{Lt_0^2}{2(1-t_0)^2}\|x_\star\|_2^2.\label{ineq:KL_1}
\end{align}
Since $X_\star$ and $Z$ are independent, we have
\begin{align*}
	p_{t_0}(y)=\int p_{X_{t_0}\mid X_\star}(y\mid x_\star)
	\,\Pdata({\rm d}x_\star).
\end{align*}
Furthermore, recognizing that the function
$r\mapsto r\log r$ is convex, Jensen's inequality yields
\begin{align*}
	\frac{p_{t_0}(y)}{p_{Y_0}(y)}\log\frac{p_{t_0}(y)}{p_{Y_0}(y)}&=\left[\int \frac{p_{X_{t_0}\mid X_\star}(y\mid x_\star)}{p_{Y_0}(y)}\,\Pdata({\rm d}x_\star)\right]\log\left[\int \frac{p_{X_{t_0}\mid X_\star}(y\mid x_\star)}{p_{Y_0}(y)}\,\Pdata({\rm d}x_\star)\right]\\
	&\leq\int \frac{p_{X_{t_0}\mid X_\star}(y\mid x_\star)}{p_{Y_0}(y)}
	\log\frac{p_{X_{t_0}\mid X_\star}(y\mid x_\star)}{p_{Y_0}(y)}
	\,\Pdata({\rm d}x_\star),
\end{align*}
or equivalently,
\[
p_{t_0}(y)\log\frac{p_{t_0}(y)}{p_{Y_0}(y)}
\leq\int p_{X_{t_0}\mid X_\star}(y\mid x_\star)
\log\frac{p_{X_{t_0}\mid X_\star}(y\mid x_\star)}{p_{Y_0}(y)}
\,\Pdata({\rm d}x_\star),
\]
In view of \eqref{eq:translated-log-density-absolute} and Assumption~\ref{ass:target}, we know that
\begin{align*}
	&\int\!\int p_{X_{t_0}\mid X_\star}(y\mid x_\star)
	\left|\log\frac{p_{X_{t_0}\mid X_\star}(y\mid x_\star)}{p_{Y_0}(y)}\right|
	\,{\rm d}y\,\Pdata({\rm d}x_\star)\\
	&\quad\leq\frac{t_0}{1-t_0}\sqrt{Ld}\,\mathbb E\bigl[\|X_\star\|_2\bigr]
	+\frac{Lt_0^2}{2(1-t_0)^2}\mathbb E\left[\|X_\star\|_2^2\right]
	<\infty.
\end{align*}
Here $\mathbb E\bigl[\|X_\star\|_2\bigr]\leq
(\mathbb E\left[\|X_\star\|_2^2\right])^{1/2}<\infty$ by
Assumption~\ref{ass:target}. In addition, we know that
\begin{align*}
	p_{t_0}(y)\log\frac{p_{t_0}(y)}{p_{Y_0}(y)} \geq -\frac{p_{Y_0}(y)}{e}
\end{align*}
since
$r\log r\geq-1/e$. Applying Fubini's theorem yields
\begin{align*}
	\mathsf{KL}(P_{t_0}\|P_{Y_0})
	&= \int p_{t_0}(y)\log\frac{p_{t_0}(y)}{p_{Y_0}(y)}\,{\rm d}y\\
	&\leq \int\left(\int p_{X_{t_0}\mid X_\star}(y\mid x_\star)
	\log\frac{p_{X_{t_0}\mid X_\star}(y\mid x_\star)}{p_{Y_0}(y)}
	\,\Pdata({\rm d}x_\star)\right){\rm d}y\\
	&= \int\left(\int p_{X_{t_0}\mid X_\star}(y\mid x_\star)
	\log\frac{p_{X_{t_0}\mid X_\star}(y\mid x_\star)}{p_{Y_0}(y)}
	\,{\rm d}y\right)\Pdata({\rm d}x_\star)\\
	&=\int\mathsf{KL}\bigl(P_{X_{t_0}\mid X_\star}(\cdot\mid x_\star)
	\|P_{Y_0}\bigr)\,\Pdata({\rm d}x_\star)\\
	&\stackrel{\eqref{ineq:KL_1}}{\leq}\int\frac{Lt_0^2}{2(1-t_0)^2}\|x_\star\|_2^2
	\,\Pdata({\rm d}x_\star)\\
	&=\frac{Lt_0^2}{2(1-t_0)^2}\mathbb E\left[\|X_\star\|_2^2\right].
\end{align*}
This proves \eqref{eq:init-KL}.

Pinsker's inequality further shows that
\begin{align*}
	\mathsf{TV}(P_{Y_0},P_{t_0})
	&=\mathsf{TV}(P_{t_0},P_{Y_0}) \leq\sqrt{\frac12\mathsf{KL}(P_{t_0}\|P_{Y_0})}
	\leq\sqrt{\frac{Lt_0^2}{4(1-t_0)^2}\mathbb E\left[\|X_\star\|_2^2\right]}
	=\frac{t_0}{2(1-t_0)}\sqrt{L\mathbb E\left[\|X_\star\|_2^2\right]},
\end{align*}
Since
$t_0\leq(N+1)^{-c_0/2}<1/2$,
$1/(2(1-t_0)) \leq1$. Consequently, by Assumptions~\ref{ass:target} and \ref{ass:schedule}, we have
\begin{align*}
	\mathsf{TV}(P_{Y_0},P_{t_0})
	&\leq t_0\sqrt{L\mathbb E\left[\|X_\star\|_2^2\right]}\\
	&\leq t_0(N+1)^{c_R/2}
	\sqrt{L\mathbb E\bigl[\|Z-\mathbb E[Z]\|_2^2\bigr]}\\
	&\leq(N+1)^{-(c_0-c_R)/2}
	\sqrt{L\mathbb E\bigl[\|Z-\mathbb E[Z]\|_2^2\bigr]}\\
	&\leq\frac{\sqrt{L\mathbb E\bigl[\|Z-\mathbb E[Z]\|_2^2\bigr]}}{(N+1)^2}.
\end{align*}
Therefore, we have finished the proof of \eqref{eq:init-absorbed}.

\subsection{Proof of Lemma~\ref{lem:source-only-regularity}}\label{app:posterior-regularity}
For every $y,z\in\mathbb R^d$, by virtue of \eqref{ineq:remainder_term}, one has
\begin{align*}
	\psource(y)\geq \psource(z)
	\exp\!\left(\frac{\|s(z)\|_2^2}{2L}\right)
	\exp\!\left(-\frac L2
	\left\|y-z-\frac{s(z)}L\right\|_2^2\right).
\end{align*}
Multiplying by $1+\|y-\mathbb E[Z]\|_2^2$ and integrating in $y$ on both sides, we obtain
\begin{align*}
	1+\mathbb E\bigl[\|Z-\mathbb E[Z]\|_2^2\bigr] \geq \psource(z)
	\exp\!\left(\frac{\|s(z)\|_2^2}{2L}\right)\int_{\mathbb R^d}(1+\|y-\mathbb E[Z]\|_2^2)
	\exp\!\left(-\frac{L}{2}\left\|y-z-\frac{s(z)}L\right\|_2^2\right)\,{\rm d}y.
\end{align*}
We let
$w=y-z-s(z)/L$. Then we have
\begin{align*}
	&\int_{\mathbb R^d}(1+\|y-\mathbb E[Z]\|_2^2)
	\exp\!\left(-\frac L2
	\left\|y-z-\frac{s(z)}L\right\|_2^2\right)\,{\rm d}y\\
	&\qquad = \int_{\mathbb R^d}\left(1+\left\|w + z + \frac{s(z)}{L} - \mathbb E[Z]\right\|_2^2\right)
	\exp\!\left(-\frac L2
	\left\|w\right\|_2^2\right)\,{\rm d}w\\
	&\qquad=\left(1+\left\|z + \frac{s(z)}{L} - \mathbb E[Z]\right\|_2^2\right)\int_{\mathbb R^d}
	\exp\!\left(-\frac L2
	\left\|w\right\|_2^2\right)\,{\rm d}w\\
	&\qquad\qquad+2\left(z + \frac{s(z)}{L} - \mathbb E[Z]\right)^\top\int_{\mathbb R^d}
	w\exp\!\left(-\frac L2
	\left\|w\right\|_2^2\right)\,{\rm d}w\\
	&\qquad\qquad+\int_{\mathbb R^d}\left\|w\right\|_2^2
	\exp\!\left(-\frac L2
	\left\|w\right\|_2^2\right)\,{\rm d}w\\
	&\qquad=\left(\frac{2\pi}{L}\right)^{d/2}
	\left(1+\left\|z+\frac{s(z)}L-\mathbb E[Z]\right\|_2^2
	+\frac dL\right),
\end{align*}
where the last equation makes use of
\begin{align*}
\int_{\mathbb R^d}e^{-L\|w\|_2^2/2}\,{\rm d}w
&=\left(\frac{2\pi}{L}\right)^{d/2},\\
\int_{\mathbb R^d}
w\exp\!\left(-\frac L2
\left\|w\right\|_2^2\right)\,{\rm d}w&=0,\\
\int_{\mathbb R^d}\|w\|_2^2e^{-L\|w\|_2^2/2}\,{\rm d}w
&=\frac dL\left(\frac{2\pi}{L}\right)^{d/2}.
\end{align*}
Therefore, we have
\begin{align*}
	&\psource(z)\exp\!\left(\frac{\|s(z)\|_2^2}{2L}\right)
	\left(1+\left\|z+\frac{s(z)}L-\mathbb E[Z]\right\|_2^2
	+\frac dL\right) \leq \left(\frac L{2\pi}\right)^{d/2}
	(1+\mathbb E\bigl[\|Z-\mathbb E[Z]\|_2^2\bigr]).
\end{align*}
Moreover, the inequality $\|v+w\|_2^2\leq2\|v\|_2^2+2\|w\|_2^2$
implies
\begin{align*}
	1+\|z-\mathbb E[Z]\|_2^2
	&\leq1+2\left\|z+\frac{s(z)}L-\mathbb E[Z]\right\|_2^2
	+\frac{2\|s(z)\|_2^2}{L^2}\\
	&\leq2\left(1+\frac{\|s(z)\|_2^2}{L^2}\right)
	\left(1+\left\|z+\frac{s(z)}L-\mathbb E[Z]\right\|_2^2
	+\frac dL\right).
\end{align*}
Combining the last two inequalities leads to
\begin{align*}
	&\psource(z)(1+\|z-\mathbb E[Z]\|_2^2)(1+\|s(z)\|_2^2)\\
	&\quad\leq2\left(\frac L{2\pi}\right)^{d/2}
	(1+\mathbb E\bigl[\|Z-\mathbb E[Z]\|_2^2\bigr])
	\left(1+\frac{\|s(z)\|_2^2}{L^2}\right)
	(1+\|s(z)\|_2^2)
	\exp\!\left(-\frac{\|s(z)\|_2^2}{2L}\right)\\
	&\quad\leq C_{d,L}(1+\|s(z)\|_2^4)
	\exp\!\left(-\frac{\|s(z)\|_2^2}{2L}\right)
	\leq C_{d,L}\left(1+\frac{16L^2}{e^2}\right),
\end{align*}
where $C_{d,L}$ is a constant depending on $L,d$ and the source moments.
Here the penultimate inequality makes use of the fact 
\begin{align*}
	\left(1+\frac{x}{L^2}\right)(1+x) = 1 + \left(1+\frac{1}{L^2}\right)x + \frac{x^2}{L^2} \leq 2\left(1+\frac{1}{L^2}\right)(1+x^2)
\end{align*}
for $x \geq 0$;
the last inequality comes from
$\sup_{u\geq0}u^2e^{-u/(2L)}=16L^2/e^2$.
As a result, we know that
\begin{equation}
	\sup_{z\in\mathbb R^d}
	\psource(z)(1+\|z-\mathbb E[Z]\|_2^2)(1+\|s(z)\|_2^2)<\infty.
	\label{eq:source-weighted-density-bound}
\end{equation}

\paragraph{Proving $p_t$ and $\mathbb E[X_\star\mid X_t=x]$ are $C^1$.} For any fixed $t\in(0,1)$, the mixture density and the posterior mean are
\begin{align*}
	p_t(x)&=\frac1{(1-t)^d}
	\int \psource\!\left(\frac{x-tx_\star}{1-t}\right)
	\,\Pdata({\rm d}x_\star),~\qquad~
	\mathbb E[X_\star\mid X_t=x]
	=\frac{\displaystyle\int x_\star
		\psource\!\left(\frac{x-tx_\star}{1-t}\right)
		\,\Pdata({\rm d}x_\star)}
	{\displaystyle\int
		\psource\!\left(\frac{x-tx_\star}{1-t}\right)
		\,\Pdata({\rm d}x_\star)}.
\end{align*}
For $\|x\|_2\leq R+1$, the identity
\[
x_\star=\frac{x-(1-t)\mathbb E[Z]}{t}
-\frac{1-t}{t}
\left(\frac{x-tx_\star}{1-t}-\mathbb E[Z]\right)
\]
implies
\begin{align*}
	1+\|x_\star\|_2&\leq
	1+\frac{R+1+(1-t)\|\mathbb E[Z]\|_2}{t}
	+\frac{1-t}{t}
	\Big\|\frac{x-tx_\star}{1-t}-\mathbb E[Z]\Big\|_2\\
	&\leq \left(1+\frac{R+1+(1-t)\|\mathbb E[Z]\|_2}{t}+\frac{1-t}{t}\right)\left(1+\Big\|\frac{x-tx_\star}{1-t}-\mathbb E[Z]\Big\|_2^2\right)\\
	&= \frac{R+2+(1-t)\|\mathbb E[Z]\|_2}{t}\left(1+\Big\|\frac{x-tx_\star}{1-t}-\mathbb E[Z]\Big\|_2^2\right).
\end{align*}
It follows from \eqref{eq:source-weighted-density-bound} that,
uniformly over $\|x\|_2\leq R+1$ and $x_\star\in\mathbb R^d$,
\begin{align*}
	&(1+\|x_\star\|_2)
	\psource\!\left(\frac{x-tx_\star}{1-t}\right)
	\left(1+\left\|s\!\left(\frac{x-tx_\star}{1-t}\right)
	\right\|_2\right)\\
	&\quad\leq C_{R,t}
	\psource\!\left(\frac{x-tx_\star}{1-t}\right)
	\left(1+\left\|\frac{x-tx_\star}{1-t}-\mathbb E[Z]
	\right\|_2^2\right)
	\left(1+\left\|s\!\left(\frac{x-tx_\star}{1-t}\right)
	\right\|_2^2\right)
	\leq C_{R,t,d,L}.
\end{align*}
Here, $C_{R,t}$ (resp.~$C_{R,t,d,L}$) is a constant depending on $R,t$ (resp.~$R,t,d,L$) and the source moments. Therefore, the numerator of the posterior mean is
absolutely integrable, and its denominator is finite and strictly
positive.

For each coordinate vector $e_j$, every $x$ with $\|x\|_2\leq R$,
and $0<|\varepsilon|<1$, the fundamental theorem of calculus reveals that
\begin{align*}
	&\frac{\psource\!\left(\frac{x+\varepsilon e_j-tx_\star}{1-t}\right)
		-\psource\!\left(\frac{x-tx_\star}{1-t}\right)}{\varepsilon}=\frac1{1-t}\int_0^1
	\psource\!\left(\frac{x+r\varepsilon e_j-tx_\star}{1-t}\right)
	s_j\!\left(\frac{x+r\varepsilon e_j-tx_\star}{1-t}\right)
	\,{\rm d}r.
\end{align*}
Applying the Dominated Convergence Theorem therefore gives
\begin{align*}
	\partial_j\int \psource\!\left(\frac{x-tx_\star}{1-t}\right)
	\,\Pdata({\rm d}x_\star)
	&=\frac1{1-t}\int
	\psource\!\left(\frac{x-tx_\star}{1-t}\right)
	s_j\!\left(\frac{x-tx_\star}{1-t}\right)
	\,\Pdata({\rm d}x_\star),\\
	\partial_j\int(x_\star)_i
	\psource\!\left(\frac{x-tx_\star}{1-t}\right)
	\,\Pdata({\rm d}x_\star)
	&=\frac1{1-t}\int(x_\star)_i
	\psource\!\left(\frac{x-tx_\star}{1-t}\right)
	s_j\!\left(\frac{x-tx_\star}{1-t}\right)
	\,\Pdata({\rm d}x_\star).
\end{align*}
Since both right-hand sides are continuous, we know that the unnormalized integrals $\int \psource\!\left(\frac{x-tx_\star}{1-t}\right)
\,\Pdata({\rm d}x_\star)$ and $\int x_\star
\psource\!\left(\frac{x-tx_\star}{1-t}\right)
\,\Pdata({\rm d}x_\star)$ are
$C^1$, and so is $p_t$. Since the denominator of the posterior
mean is positive, the quotient rule shows that the posterior
mean is $C^1$ as well.

\paragraph{Proving \eqref{eq:posterior-normalization-differentiation} and \eqref{eq:posterior-mean-differentiation}.}
By Bayes' formula and the quotient rule, we have
\begin{align*}
	p_{X_\star\mid X_t}(x_\star\mid x)
	&=\frac{\psource\!\left(\frac{x-tx_\star}{1-t}\right)}
	{(1-t)^d p_t(x)},\\
	\nabla_xp_{X_\star\mid X_t}(x_\star\mid x)
	&=\frac{\psource\!\left(\frac{x-tx_\star}{1-t}\right)}
	{(1-t)^{d+1}p_t(x)}
	s\!\left(\frac{x-tx_\star}{1-t}\right)
	-\frac{\psource\!\left(\frac{x-tx_\star}{1-t}\right)}
	{(1-t)^d p_t(x)^2}\nabla p_t(x).
\end{align*}
For every fixed $x$, the domination established above shows that
both terms in the second equation are absolutely integrable with
respect to $\Pdata$, even after multiplication by $1+\|x_\star\|_2$.
Consequently, the derivative formula for the unnormalized density implies
\begin{align*}
	\int\nabla_xp_{X_\star\mid X_t}(x_\star\mid x)
	\,\Pdata({\rm d}x_\star)
	&=\frac{\nabla p_t(x)}{p_t(x)}
	-\frac{\nabla p_t(x)}{p_t(x)}=0\\
	&=\nabla_x1
	=\nabla_x\int p_{X_\star\mid X_t}(x_\star\mid x)
	\,\Pdata({\rm d}x_\star),
\end{align*}
which proves \eqref{eq:posterior-normalization-differentiation}.
Moreover, applying the quotient rule to the posterior mean and using
the derivative formulas for the unnormalized integrals, we obtain
\begin{align*}
	\nabla_x\mathbb E[X_\star\mid X_t=x]
	&=\nabla_x\left[\frac{\int x_\star
		\psource\!\left(\frac{x-tx_\star}{1-t}\right)
		\,\Pdata({\rm d}x_\star)}
	{(1-t)^{d}p_t(x)}\right]\\
	&=\frac{1}{(1-t)^{d+1}p_t(x)}
	\int x_\star \psource\!\left(\frac{x-tx_\star}{1-t}\right)
	\left[s\!\left(\frac{x-tx_\star}{1-t}\right)\right]^\top
	\,\Pdata({\rm d}x_\star)\\
	&\qquad-\frac{1}{(1-t)^d p_t(x)^2}
	\left[\int x_\star \psource\!\left(\frac{x-tx_\star}{1-t}\right)
	\,\Pdata({\rm d}x_\star)\right][\nabla p_t(x)]^\top\\
	&=\int x_\star
	\left[\nabla_xp_{X_\star\mid X_t}(x_\star\mid x)\right]^\top
	\,\Pdata({\rm d}x_\star)\\
	&=\int (x_\star-x)
	\left[\nabla_xp_{X_\star\mid X_t}(x_\star\mid x)\right]^\top
	\,\Pdata({\rm d}x_\star)+x\left[\int\nabla_xp_{X_\star\mid X_t}(x_\star\mid x)
	\,\Pdata({\rm d}x_\star)\right]^\top\\
	&=\int (x_\star-x)
	\left[\nabla_xp_{X_\star\mid X_t}(x_\star\mid x)\right]^\top
	\,\Pdata({\rm d}x_\star).
\end{align*}
Here the last equation comes from
\eqref{eq:posterior-normalization-differentiation}.
This proves \eqref{eq:posterior-mean-differentiation}, and hence
$v_t(x)=(\mathbb E[X_\star\mid X_t=x]-x)/(1-t)$ is $C^1$.

\paragraph{Proving continuity of the conditional second moments.}
Since $Z=(x-tX_\star)/(1-t)$ when $X_t=x$, Bayes' formula gives
\begin{align*}
	\mathbb E[\|Z-\mathbb E[Z]\|_2^2\mid X_t=x]
	&=\frac{\displaystyle\int
		\left\|\frac{x-tx_\star}{1-t}-\mathbb E[Z]\right\|_2^2
		\psource\!\left(\frac{x-tx_\star}{1-t}\right)
		\,\Pdata({\rm d}x_\star)}
	{\displaystyle\int
		\psource\!\left(\frac{x-tx_\star}{1-t}\right)
		\,\Pdata({\rm d}x_\star)},\\
	\mathbb E[\|s(Z)\|_2^2\mid X_t=x]
	&=\frac{\displaystyle\int
		\left\|s\!\left(\frac{x-tx_\star}{1-t}\right)\right\|_2^2
		\psource\!\left(\frac{x-tx_\star}{1-t}\right)
		\,\Pdata({\rm d}x_\star)}
	{\displaystyle\int
		\psource\!\left(\frac{x-tx_\star}{1-t}\right)
		\,\Pdata({\rm d}x_\star)}.
\end{align*}
By virtue of \eqref{eq:source-weighted-density-bound}, both numerator
integrands are bounded uniformly over $x,x_\star\in\mathbb R^d$.
Since $\psource$ and $s$ are continuous,
the Dominated Convergence Theorem shows that both numerators are finite
and continuous. Their common denominator is positive and continuous,
so both conditional second moments are finite and continuous as well.
Therefore, $T_n(x)=x+(t_n-t_{n-1})v_{t_{n-1}}(x)$ is $C^1$,
and $G_n$ is closed. In particular, $G_n$ is Borel.

\subsection{Proof of Lemma~\ref{lem:velocity-jacobian}}\label{app:velocity-jacobian}
\begin{proof}
Recall from \eqref{eq:conditional-velocity} and \eqref{eq:marginal-velocity} that
\begin{align*}
 v_t(x)
 &=\mathbb E\left[v_t(x|X_\star)\mymid X_t=x\right]
 =\mathbb E\left[\frac{X_\star-x}{1-t}\mymid X_t=x\right]\\
 &=\int_{x_\star}p_{X_\star\mymid X_t}(x_\star\mymid x)
       \frac{x_\star-x}{1-t}\,\Pdata({\rm d}x_{\star}).
\end{align*}
Consequently, by \eqref{eq:posterior-mean-differentiation} in
Lemma~\ref{lem:source-only-regularity}, one has
\begin{align*}
 \nabla_xv_t(x)
 &=\frac{1}{1-t}
   \left(\nabla_x\mathbb E[X_\star\mymid X_t=x]-I\right)\\
 &=-\frac{1}{1-t}I+\frac{1}{1-t}\int_{x_\star}
   x_\star\left[\nabla_xp_{X_\star\mymid X_t}
                    (x_\star\mymid x)\right]^\top\Pdata({\rm d}x_{\star})\\
 &=-\frac{1}{1-t}I+\frac{1}{1-t}\int_{x_\star}
   (x_\star-x)\left[\nabla_xp_{X_\star\mymid X_t}
                    (x_\star\mymid x)\right]^\top\Pdata({\rm d}x_{\star}).
\end{align*}
For any fixed $x_\star$, Bayes' rule and the fact that
$\nabla_xp(x)=p(x)\nabla_x\log p(x)$ imply that
\begin{align}\label{eq:posterior-density-gradient}
 \nabla_xp_{X_\star\mymid X_t}(x_\star\mymid x)
 &=p_{X_\star\mymid X_t}(x_\star\mymid x)
   \nabla_x\log p_{X_\star\mymid X_t}(x_\star\mymid x)\notag\\
 &=p_{X_\star\mymid X_t}(x_\star\mymid x)
   \nabla_x\log\left(
   \frac{p_{X_t\mymid X_\star}(x\mymid x_\star)}{p_t(x)}
   \right)\notag\\
 &=p_{X_\star\mymid X_t}(x_\star\mymid x)
   \left[\nabla_x\log p_{X_t\mymid X_\star}(x\mymid x_\star)
          -\nabla_x\log p_t(x)\right].
\end{align}
Putting the previous two identities together, we obtain
\begin{equation}\label{eq:velocity-jacobian-integral}
\begin{aligned}
 \nabla_xv_t(x)
 &=-\frac{1}{1-t}I+\frac{1}{1-t}\int_{x_\star}
   p_{X_\star\mymid X_t}(x_\star\mymid x)(x_\star-x)
   \left[\nabla_x\log p_{X_t\mymid X_\star}(x\mymid x_\star)
          -\nabla_x\log p_t(x)\right]^\top\Pdata({\rm d}x_{\star}).
\end{aligned}
\end{equation}
In addition, \eqref{eq:posterior-density-gradient} and
\eqref{eq:posterior-normalization-differentiation} in
Lemma~\ref{lem:source-only-regularity} together show that for any fixed
$y\in\mathbb R^d$,
\begin{align}
 &\int_{x_\star}p_{X_\star\mymid X_t}(x_\star\mymid x)y
   \left[\nabla_x\log p_{X_t\mymid X_\star}(x\mymid x_\star)
          -\nabla_x\log p_t(x)\right]^\top\Pdata({\rm d}x_{\star})\notag\\
 &\quad=\int_{x_\star}y
   \left[\nabla_xp_{X_\star\mymid X_t}
          (x_\star\mymid x)\right]^\top\Pdata({\rm d}x_{\star})\notag\\
 &\quad=y\left[\nabla_x\int_{x_\star}
   p_{X_\star\mymid X_t}(x_\star\mymid x)\,\Pdata({\rm d}x_{\star})
   \right]^\top\notag\\
 &\quad=y(\nabla_x1)^\top=0.\label{eq:centered-posterior-score}
\end{align}

Setting $y = \mathbb E[X_\star\mymid X_t=x]-x$ in \eqref{eq:centered-posterior-score} yields
\begin{align*}
 &\int_{x_\star}p_{X_\star\mymid X_t}(x_\star\mymid x)
   (x_\star-x)
   \left[\nabla_x\log p_{X_t\mymid X_\star}(x\mymid x_\star)
          -\nabla_x\log p_t(x)\right]^\top\Pdata({\rm d}x_{\star})\\
 &=\int_{x_\star}p_{X_\star\mymid X_t}(x_\star\mymid x)
   \bigl(x_\star-\mathbb E[X_\star\mymid X_t=x]\bigr)
   \left[\nabla_x\log p_{X_t\mymid X_\star}(x\mymid x_\star)
          -\nabla_x\log p_t(x)\right]^\top\Pdata({\rm d}x_{\star})\\
 &\quad+\int_{x_\star}p_{X_\star\mymid X_t}(x_\star\mymid x)
   \bigl(\mathbb E[X_\star\mymid X_t=x]-x\bigr)
   \left[\nabla_x\log p_{X_t\mymid X_\star}(x\mymid x_\star)
          -\nabla_x\log p_t(x)\right]^\top\Pdata({\rm d}x_{\star})\\
 &=
   \int_{x_\star}p_{X_\star\mymid X_t}(x_\star\mymid x)
   \bigl(x_\star-\mathbb E[X_\star\mymid X_t=x]\bigr)
   \left[\nabla_x\log p_{X_t\mymid X_\star}(x\mymid x_\star)
          -\nabla_x\log p_t(x)\right]^\top\Pdata({\rm d}x_{\star})\\
 &=\int_{x_\star}p_{X_\star\mymid X_t}(x_\star\mymid x)
   \bigl(x_\star-\mathbb E[X_\star\mymid X_t=x]\bigr)
   \left[\nabla_x\log p_{X_t\mymid X_\star}
          (x\mymid x_\star)\right]^\top\Pdata({\rm d}x_{\star})\\
 &\quad-\left[\int_{x_\star}
   p_{X_\star\mymid X_t}(x_\star\mymid x)
   \bigl(x_\star-\mathbb E[X_\star\mymid X_t=x]\bigr)
   \,\Pdata({\rm d}x_{\star})\right]\left[\nabla_x\log p_t(x)\right]^\top\\
 &=\int_{x_\star}p_{X_\star\mymid X_t}(x_\star\mymid x)
   \bigl(x_\star-\mathbb E[X_\star\mymid X_t=x]\bigr)
   \left[\nabla_x\log p_{X_t\mymid X_\star}
          (x\mymid x_\star)\right]^\top\Pdata({\rm d}x_{\star})\\
 &\quad-\mathbb E\left[
   X_\star-\mathbb E[X_\star\mymid X_t=x]\mymid X_t=x\right]
   \left[\nabla_x\log p_t(x)\right]^\top\\
 &=\int_{x_\star}p_{X_\star\mymid X_t}(x_\star\mymid x)
   \bigl(x_\star-\mathbb E[X_\star\mymid X_t=x]\bigr)
   \left[\nabla_x\log p_{X_t\mymid X_\star}
          (x\mymid x_\star)\right]^\top\Pdata({\rm d}x_{\star}).
\end{align*}
Equation \eqref{eq:conditional-density} together with the chain rule implies
\begin{align}
 \nabla_x\log p_{X_t\mymid X_\star}(x\mymid x_\star)
 &=\nabla_x\log\left((1-t)^{-d}
   \psource\!\left(\frac{x-tx_\star}{1-t}\right)\right)\notag\\
 &=\nabla_x\log \psource\!\left(\frac{x-tx_\star}{1-t}\right)\notag\\
 &=\left(\nabla_x\frac{x-tx_\star}{1-t}\right)^\top
   \nabla_z\log \psource\!\left(\frac{x-tx_\star}{1-t}\right)\notag\\
 &=\frac{1}{1-t}
   \nabla_z\log \psource\!\left(\frac{x-tx_\star}{1-t}\right).
\end{align}
Combining the previous two equations together, we have
\begin{align}\label{eq:posterior-score-covariance}
 &\int_{x_\star}p_{X_\star\mymid X_t}(x_\star\mymid x)
   (x_\star-x)
   \left[\nabla_x\log p_{X_t\mymid X_\star}(x\mymid x_\star)
          -\nabla_x\log p_t(x)\right]^\top\Pdata({\rm d}x_{\star})\notag\\
 &\quad=\frac{1}{1-t}\int_{x_\star}
   p_{X_\star\mymid X_t}(x_\star\mymid x)
   \bigl(x_\star-\mathbb E[X_\star\mymid X_t=x]\bigr)
   \left[\nabla_z\log \psource\!\left(
   \frac{x-tx_\star}{1-t}\right)\right]^\top\Pdata({\rm d}x_{\star})\notag\\
 &\quad=\frac{1}{1-t}\int_{x_\star}
   p_{X_\star\mymid X_t}(x_\star\mymid x)
   \bigl(x_\star-\mathbb E[X_\star\mymid X_t=x]\bigr)\notag\\
 &\qquad\qquad\times\left[
   \nabla_z\log \psource\!\left(\frac{x-tx_\star}{1-t}\right)
   -\nabla_z\log \psource\!\left(
   \frac{x-t\mathbb E[X_\star\mymid X_t=x]}{1-t}\right)
   \right]^\top\Pdata({\rm d}x_{\star})\notag\\
 &\qquad+\frac{1}{1-t}\int_{x_\star}
   p_{X_\star\mymid X_t}(x_\star\mymid x)
   \bigl(x_\star-\mathbb E[X_\star\mymid X_t=x]\bigr)\left[
   \nabla_z\log \psource\!\left(
   \frac{x-t\mathbb E[X_\star\mymid X_t=x]}{1-t}\right)
   \right]^\top\Pdata({\rm d}x_{\star})\notag\\
 &\quad=\frac{1}{1-t}\int_{x_\star}
   p_{X_\star\mymid X_t}(x_\star\mymid x)
   \bigl(x_\star-\mathbb E[X_\star\mymid X_t=x]\bigr)\notag\\
 &\qquad\qquad\times\left[
   \nabla_z\log \psource\!\left(\frac{x-tx_\star}{1-t}\right)
   -\nabla_z\log \psource\!\left(
   \frac{x-t\mathbb E[X_\star\mymid X_t=x]}{1-t}\right)
   \right]^\top\Pdata({\rm d}x_{\star})\notag\\
 &\qquad+\frac{1}{1-t}\mathbb E\left[
   X_\star-\mathbb E[X_\star\mymid X_t=x]\mymid X_t=x\right]\left[
   \nabla_z\log \psource\!\left(
   \frac{x-t\mathbb E[X_\star\mymid X_t=x]}{1-t}\right)
   \right]^\top\notag\\
 &\quad=\frac{1}{1-t}\int_{x_\star}
   p_{X_\star\mymid X_t}(x_\star\mymid x)
   \bigl(x_\star-\mathbb E[X_\star\mymid X_t=x]\bigr)\left[
   s\left(\frac{x-tx_\star}{1-t}\right)
   -s\left(
   \frac{x-t\mathbb E[X_\star\mymid X_t=x]}{1-t}\right)
   \right]^\top\Pdata({\rm d}x_{\star})\notag\\
   &\quad= \frac{1}{1-t}\mathbb{E}\left[\bigl(X_\star-\mathbb E[X_\star\mymid X_t=x]\bigr)\left[
   s\left(\frac{x-tX_\star}{1-t}\right)
   -s\left(
   \frac{x-t\mathbb E[X_\star\mymid X_t=x]}{1-t}\right)
   \right]^\top~\bigg|~X_t = x\right].
\end{align}
Conditional on $X_t=x$, one has
\begin{align*}
	t(X_\star-\mathbb E[X_\star\mid X_t=x])
	=-(1-t)(Z-\mathbb E[Z\mid X_t=x]).
\end{align*}
Combining \eqref{eq:velocity-jacobian-integral}, \eqref{eq:posterior-score-covariance} and the previous equation, one has
\begin{align}
	\nabla_xv_t(x)
	&=-\frac{1}{1-t}I-\frac{1}{t(1-t)}\mathbb{E}\left[(Z-\mathbb E[Z\mid X_t=x])\left[
	s\left(Z\right)
	-s\left(
	\mathbb E[Z\mymid X_t=x]\right)
	\right]^\top~\big|~X_t = x\right].\label{eq:gradient_v_t}
\end{align}
Since $s(\cdot)$ is $L$-Lipschitz, one has
\begin{align}
	\left\|\nabla v_t(x)+\frac{I_d}{1-t}\right\|_{\mathrm{op}}
	&\le
	\left\|\nabla v_t(x)+\frac{I_d}{1-t}\right\|_{\mathrm F}\notag\\
	&\le\frac{1}{t(1-t)}
	\mathbb E\left[\|Z-\mathbb E[Z\mid X_t=x]\|_2\left\|s\left(Z\right)
	-s\left(\mathbb E[Z\mymid X_t=x]\right)\right\|_2\,\middle|\,X_t=x\right]\notag\\
	&\le\frac{L}{t(1-t)}
	\mathbb E\left[\|Z-\mathbb E[Z\mid X_t=x]\|_2^2
	\,\middle|\,X_t=x\right].
	\label{eq:jacobian-lipschitz}
\end{align}
Moreover, the tower law and \eqref{eq:gradient_v_t} also imply that
\begin{align*}
	\nabla_xv_t(x)
	&=-\frac{1}{1-t}I-\frac{1}{t(1-t)}\mathbb{E}\left[(Z-\mathbb E[Z\mid X_t=x])s\left(Z\right)^\top~\big|~X_t = x\right].
\end{align*}
The Cauchy--Schwarz inequality tells us that
\begin{align*}
	\left\|\nabla v_t(x)+\frac{I_d}{1-t}\right\|_{\mathrm F}^2&\le\frac{
		\mathbb E[\|Z-\mathbb E[Z\mid X_t=x]\|_2^2\mid X_t=x]\,
		\mathbb E[\|s(Z)\|_2^2\mid X_t=x]}
	{t^2(1-t)^2}\\
	&\le\frac{
		\mathbb E[\|Z-\mathbb E[Z]\|_2^2\mid X_t=x]\,
		\mathbb E[\|s(Z)\|_2^2\mid X_t=x]}
	{t^2(1-t)^2}.
\end{align*}
Then the basic inequality $\|A+B\|_{\rm F}^2 \leq 2(\|A\|_{\rm F}^2+\|B\|_{\rm F}^2)$ implies \eqref{eq:jacobian-frobenius}.
\end{proof}

\subsection{Proof of Lemma~\ref{lem:restriction}}\label{app:restriction-proof}

\begin{proof}
We fix $1\leq n\leq N$. In view of Lemma~\ref{lem:source-only-regularity}, we know that
$T_n$ is $C^1$ and $G_n$ is Borel.

\paragraph{Proving injectivity on $G_n$.}
We prove injectivity by contradiction. Suppose that $x,y\in G_n$ satisfy $T_n(x)=T_n(y)$, and let
$\alpha=(y-x)/(1-t_{n-1})$. Since
\begin{align*}
 T_n(x)
 &=x+\frac{t_n-t_{n-1}}{t_{n-1}}
       \bigl(x-\mathbb E[Z\mid X_{t_{n-1}}=x]\bigr),
\end{align*}
we know that
\begin{equation}
 \mathbb E[Z\mid X_{t_{n-1}}=y]
 -\mathbb E[Z\mid X_{t_{n-1}}=x]
 =\frac{t_n}{t_n-t_{n-1}}(y-x) = \frac{t_n(1-t_{n-1})}{t_n-t_{n-1}}\alpha.
 \label{eq:collision-conditional-means}
\end{equation}
By Jensen's inequality and the definition of $G_n$ in
\eqref{eq:good-set}, we have
\begin{align}
 \left\|\alpha\right\|_2
 &\leq\frac{t_n-t_{n-1}}{(1-t_{n-1})t_n}
 \left(\|\mathbb E[Z\mid X_{t_{n-1}}=y]-\mathbb E[Z]\|_2
       +\|\mathbb E[Z\mid X_{t_{n-1}}=x]-\mathbb E[Z]\|_2\right)
 \notag\\
 &\leq\frac{t_n-t_{n-1}}{(1-t_{n-1})t_n}
 \left\{\sqrt{\mathbb E[\|Z-\mathbb E[Z]\|_2^2\mid X_{t_{n-1}}=y]}
 +\sqrt{\mathbb E[\|Z-\mathbb E[Z]\|_2^2\mid X_{t_{n-1}}=x]}\right\}
 \notag\\
 &\leq\frac{t_{n-1}(1-t_n)}
 {8(1-t_{n-1})t_n\sqrt{L\{d+12\log(N+1)\}}}
 \leq\frac1{8\sqrt{L\{d+12\log(N+1)\}}}.
 \label{eq:collision-distance}
\end{align}
Here the last inequality arises from
$(1-t_{n-1})t_n=t_{n-1}(1-t_n)+(t_n-t_{n-1}) \geq t_{n-1}(1-t_n)$.

Let $P$ and $Q$ be the posterior laws of $X_\star$ given
$X_{t_{n-1}}=x$ and $X_{t_{n-1}}=y$, respectively. We let
\begin{align}
 \ell(x_\star)
 :=\log\frac{{\rm d}Q}{{\rm d}P}(x_\star)
 &=\log \psource\!\left(\frac{y-t_{n-1}x_\star}{1-t_{n-1}}\right)
   -\log \psource\!\left(\frac{x-t_{n-1}x_\star}{1-t_{n-1}}\right)
   +\log\frac{p_{t_{n-1}}(x)}{p_{t_{n-1}}(y)}.
 \label{eq:posterior-log-ratio}
\end{align}
Applying \eqref{ineq:remainder_term} at the two endpoints reveals that
\begin{subequations}
	\begin{align}
		\left|\ell(x_\star)-\log\frac{p_{t_{n-1}}(x)}{p_{t_{n-1}}(y)}
		-\alpha^\top s\!\left(\frac{x-t_{n-1}x_\star}{1-t_{n-1}}\right)\right|
		&\leq\frac L2\left\|\frac{y-x}{1-t_{n-1}}\right\|_2^2,\label{eq:taylor_remainder1}\\
		\left|\ell(x_\star)-\log\frac{p_{t_{n-1}}(x)}{p_{t_{n-1}}(y)}
		-\alpha^\top s\!\left(\frac{y-t_{n-1}x_\star}{1-t_{n-1}}\right)\right|
		&\leq\frac L2\left\|\frac{y-x}{1-t_{n-1}}\right\|_2^2.\label{eq:taylor_remainder2}
	\end{align}
\end{subequations}
Moreover, by the definition of $G_n$, we have
\begin{subequations}
	\begin{align}
		\mathbb E_P\!\left[\left\|s\!\left(
		\frac{x-t_{n-1}X_\star}{1-t_{n-1}}\right)\right\|_2^2\right]
		&= \mathbb{E}\left[\|s(Z)\|_2^2\mid X_{t_{n-1}}=x\right] \leq 4L\{d+12\log(N+1)\},\label{eq:score_square_1}\\
		\mathbb E_Q\!\left[\left\|s\!\left(
		\frac{y-t_{n-1}X_\star}{1-t_{n-1}}\right)\right\|_2^2\right]
		&= \mathbb{E}\left[\|s(Z)\|_2^2\mid X_{t_{n-1}}=y\right] \leq 4L\{d+12\log(N+1)\}\label{eq:score_square_2}.
	\end{align}
\end{subequations}
Thus $\ell(X_\star)$ has finite second moments under both laws.
In particular, both relative entropies are finite, and
\begin{align}
 0\leq\mathsf{KL}(P\|Q)+\mathsf{KL}(Q\|P)
 &=\mathbb E_Q[\ell(X_\star)]-\mathbb E_P[\ell(X_\star)]\notag\\
 &=\mathbb E_Q\left[\ell(X_\star)-\log\frac{p_{t_{n-1}}(x)}{p_{t_{n-1}}(y)}\right]-\mathbb E_P\left[\ell(X_\star)-\log\frac{p_{t_{n-1}}(x)}{p_{t_{n-1}}(y)}\right]\notag\\
 &\stackrel{\eqref{eq:taylor_remainder1}-\eqref{eq:score_square_2}}{\leq}4\sqrt{L\{d+12\log(N+1)\}}\left\|\alpha\right\|_2+L\left\|\alpha\right\|_2^2\notag\\
 &\stackrel{\eqref{eq:collision-distance}}{\leq}
 \frac12+\frac1{64\{d+12\log(N+1)\}}
 \leq\frac9{16}.
 \label{eq:posterior-sym-kl}
\end{align}
The same two Taylor inequalities also allow us to control the
variances. More specifically, using $(u+v)^2\leq2u^2+2v^2$, we obtain
\begin{align}
 &\operatorname{Var}_P(\ell(X_\star))
  +\operatorname{Var}_Q(\ell(X_\star))\notag\\
 &\quad\leq
 \mathbb E_P\!\left[\left(\ell(X_\star)
       -\log\frac{p_{t_{n-1}}(x)}{p_{t_{n-1}}(y)}\right)^2\right]
 +\mathbb E_Q\!\left[\left(\ell(X_\star)
       -\log\frac{p_{t_{n-1}}(x)}{p_{t_{n-1}}(y)}\right)^2\right]\notag\\
 &\quad\leq2\left\|\alpha\right\|_2^2\left\{
 \mathbb E_P\!\left[\left\|s\!\left(
 \frac{x-t_{n-1}X_\star}{1-t_{n-1}}\right)\right\|_2^2\right]
 +\mathbb E_Q\!\left[\left\|s\!\left(
 \frac{y-t_{n-1}X_\star}{1-t_{n-1}}\right)\right\|_2^2\right]
 \right\}+L^2\left\|\alpha\right\|_2^4\notag\\
 &\quad\leq16L\{d+12\log(N+1)\}\left\|\alpha\right\|_2^2+L^2\left\|\alpha\right\|_2^4.
 \label{eq:log-ratio-variance}
\end{align}

We next compare expectations under $P$ and $Q$. Since
$(\exp(u)-1)u\geq2(\exp(u)-1)^2/(\exp(u)+1)$ for every $u\in\mathbb R$ and
\begin{align}\label{eq:relation_ell}
	\exp(\ell)\,{\rm d}P={\rm d}Q,
\end{align}
we have
\begin{align}\label{eq:posterior-divergence-comparison}
 \int\frac{(\exp(\ell)-1)^2}{\exp(\ell)+1}\,{\rm d}P
 &\leq\frac12\int(\exp(\ell)-1)\ell\,{\rm d}P
 =\frac12\int(\frac{{\rm d}Q}{{\rm d}P}-1)\log\frac{{\rm d}Q}{{\rm d}P}\,{\rm d}P=\frac{\mathsf{KL}(P\|Q)+\mathsf{KL}(Q\|P)}2.
\end{align}
For any real-valued $f$ with finite second moments under both laws,
we have
\begin{align}\label{eq:posterior-centered-difference}
 &\mathbb E_Q[f(X_\star)]-\mathbb E_P[f(X_\star)]\notag\\
 &\qquad= \mathbb E_Q\left[f(X_\star)-\frac{\mathbb E_P[f(X_\star)]+
 	\mathbb E_Q[f(X_\star)]}{2}\right]-\mathbb E_P\left[f(X_\star)-\frac{\mathbb E_P[f(X_\star)]+
 	\mathbb E_Q[f(X_\star)]}{2}\right]\notag\\
 &\qquad\stackrel{\eqref{eq:relation_ell}}{=}\int\left(f-\frac{\mathbb E_P[f(X_\star)]+
                  \mathbb E_Q[f(X_\star)]}{2}\right)
                  (\exp(\ell)-1)\,{\rm d}P.
\end{align}
By virtue of the Cauchy--Schwarz inequality, \eqref{eq:posterior-divergence-comparison} and \eqref{eq:posterior-centered-difference}, it follows that
\begin{align*}
 &\left|\mathbb E_Q[f(X_\star)]-\mathbb E_P[f(X_\star)]\right|^2\\
 &\quad\leq
 \left[\int\left(f-\frac{\mathbb E_P[f(X_\star)]+
                  \mathbb E_Q[f(X_\star)]}{2}\right)^2
                  (\exp(\ell)+1)\,{\rm d}P\right]
 \left[\int\frac{(\exp(\ell)-1)^2}{\exp(\ell)+1}\,{\rm d}P\right]\\
 &\quad\leq\frac{\mathsf{KL}(P\|Q)+\mathsf{KL}(Q\|P)}2
 \Biggl\{\mathbb E_P\left[\left(f(X_{\star})-\frac{\mathbb E_P[f(X_\star)]+
 	\mathbb E_Q[f(X_\star)]}{2}\right)^2\right]\\
 &\qquad{}+\mathbb E_Q\left[\left(f(X_{\star})-\frac{\mathbb E_P[f(X_\star)]+
 	\mathbb E_Q[f(X_\star)]}{2}\right)^2\right]\Biggr\}\\
 &\quad=\frac{\mathsf{KL}(P\|Q)+\mathsf{KL}(Q\|P)}2
 \left\{\operatorname{Var}_P(f(X_\star))
       +\operatorname{Var}_Q(f(X_\star))
       +\frac12\left|\mathbb E_Q[f(X_\star)]-
                     \mathbb E_P[f(X_\star)]\right|^2\right\}.
\end{align*}
Rearranging the terms and making use of \eqref{eq:posterior-sym-kl} yields
\begin{align}
 &\left|\mathbb E_Q[f(X_\star)]-\mathbb E_P[f(X_\star)]\right|^2
 \leq\{\mathsf{KL}(P\|Q)+\mathsf{KL}(Q\|P)\}
 \{\operatorname{Var}_P(f(X_\star))+
   \operatorname{Var}_Q(f(X_\star))\}.
 \label{eq:restricted-posterior-mean-comparison}
\end{align}
By letting $f=\ell$ and using
$\mathbb E_Q[\ell(X_\star)]-\mathbb E_P[\ell(X_\star)]
=\mathsf{KL}(P\|Q)+\mathsf{KL}(Q\|P)$, we obtain
\begin{align}\label{eq:symmetric-kl-variance-bound}
 \mathsf{KL}(P\|Q)+\mathsf{KL}(Q\|P)
 \leq\operatorname{Var}_P(\ell(X_\star))
     +\operatorname{Var}_Q(\ell(X_\star)).
\end{align}
Putting \eqref{eq:log-ratio-variance} and \eqref{eq:symmetric-kl-variance-bound} together, one has
\begin{align}
	&\left|\mathbb E_Q[f(X_\star)]-\mathbb E_P[f(X_\star)]\right|^2\notag\\ &\qquad\leq \left(16L\{d+12\log(N+1)\}\left\|\alpha\right\|_2^2+L^2\left\|\alpha\right\|_2^4\right)\{\operatorname{Var}_P(f(X_\star))+
	\operatorname{Var}_Q(f(X_\star))\}.\label{eq:posterior-expectation-difference}
\end{align}

For every unit vector $v\in\mathbb R^d$, we let
\[
 f(x_\star)=v^\top\frac{x-t_{n-1}x_\star}{1-t_{n-1}}.
\]
Under $P$, the vector $(x-t_{n-1}X_\star)/(1-t_{n-1})$ has the
conditional law of $Z$ given $X_{t_{n-1}}=x$. Under $Q$, 
\begin{align*}
	\frac{y-t_{n-1}X_\star}{1-t_{n-1}} = \frac{x-t_{n-1}X_\star}{1-t_{n-1}} + \frac{y-x}{1-t_{n-1}}
\end{align*}
has the conditional law of $Z$ given
$X_{t_{n-1}}=y$. By the definition of $G_n$, we have
\begin{align}
 &\operatorname{Var}_P(f(X_\star))+
   \operatorname{Var}_Q(f(X_\star))\notag\\
   &\quad=\Var[v^\top Z\mid X_{t_{n-1}}=x]+\Var\left[v^\top\left(Z-\frac{y-x}{1-t_{n-1}}\right)~\Big|~ X_{t_{n-1}}=y\right]\notag\\
   &\quad=\mathbb E\bigl[(v^\top(Z-\mathbb E[Z\mid X_{t_{n-1}}=x]))^2
   \mid X_{t_{n-1}}=x\bigr]
   +\mathbb E\bigl[(v^\top(Z-\mathbb E[Z\mid X_{t_{n-1}}=y]))^2
   \mid X_{t_{n-1}}=y\bigr]\notag\\
 &\quad\leq
 \mathbb E[\|Z-\mathbb E[Z\mid X_{t_{n-1}}=x]\|_2^2
            \mid X_{t_{n-1}}=x]
 +\mathbb E[\|Z-\mathbb E[Z\mid X_{t_{n-1}}=y]\|_2^2
            \mid X_{t_{n-1}}=y]\notag\\
 &\quad\leq
 \frac{t_{n-1}^2(1-t_n)^2}
 {128L(t_n-t_{n-1})^2\{d+12\log(N+1)\}}.\label{eq:posterior-directional-variances}
\end{align}
Taking the supremum over unit vectors $v$ yields
\begin{align}
 &\left\|\mathbb E_Q\!\left[
 \frac{x-t_{n-1}X_\star}{1-t_{n-1}}\right]
 -\mathbb E_P\!\left[
 \frac{x-t_{n-1}X_\star}{1-t_{n-1}}\right]\right\|_2^2\notag\\
 &\qquad=\sup_{v:\|v\|_2 = 1}\left[v^\top\left(\mathbb E_Q\!\left[
 \frac{x-t_{n-1}X_\star}{1-t_{n-1}}\right]
 -\mathbb E_P\!\left[
 \frac{x-t_{n-1}X_\star}{1-t_{n-1}}\right]\right)\right]^2\notag\\
 &\qquad=\sup_{v:\|v\|_2 = 1}\left[\mathbb E_Q\!\left[
 v^\top\frac{x-t_{n-1}X_\star}{1-t_{n-1}}\right]
 -\mathbb E_P\!\left[
 v^\top\frac{x-t_{n-1}X_\star}{1-t_{n-1}}\right]\right]^2\notag\\
 &\qquad\stackrel{\eqref{eq:posterior-expectation-difference}~\text{and}~\eqref{eq:posterior-directional-variances}}{\leq}\frac{t_{n-1}^2(1-t_n)^2}
 {128L(t_n-t_{n-1})^2\{d+12\log(N+1)\}}
 \left(16L\{d+12\log(N+1)\}\left\|\alpha\right\|_2^2+L^2\left\|\alpha\right\|_2^4\right).
 \label{eq:collision-mean-bound}
\end{align}
On the other hand, \eqref{eq:collision-conditional-means} gives
\begin{align*}
 \mathbb E_Q\!\left[
 \frac{x-t_{n-1}X_\star}{1-t_{n-1}}\right]
 -\mathbb E_P\!\left[
 \frac{x-t_{n-1}X_\star}{1-t_{n-1}}\right]
 &=\mathbb E[Z\mid X_{t_{n-1}}=y]
         -\mathbb E[Z\mid X_{t_{n-1}}=x]-\alpha\\
 &=\left(\frac{(1-t_{n-1})t_n}{t_n-t_{n-1}}-1\right)\alpha
 =\frac{t_{n-1}(1-t_n)}{t_n-t_{n-1}}\alpha.
\end{align*}
If $\alpha\ne0$, substituting this identity into
\eqref{eq:collision-mean-bound} and cancelling the positive
factor $t_{n-1}^2(1-t_n)^2\|\alpha\|_2^2/(t_n-t_{n-1})^2$ gives
\begin{align*}
 1&\leq\frac18+
 \frac{L\|\alpha\|_2^2}{128\{d+12\log(N+1)\}}
 \stackrel{\eqref{eq:collision-distance}}{\leq}
 \frac18+\frac1{8192\{d+12\log(N+1)\}^2}<1,
\end{align*}
a contradiction! Therefore, $\alpha=0$ and $x=y$, which proves
injectivity on $G_n$.

\paragraph{Controlling the Jacobian on $G_n$.}
For every $x\in G_n$, \eqref{eq:jacobian-cs} and
\eqref{eq:good-set} imply that
\begin{align*}
 &\left\|\nabla_xv_{t_{n-1}}(x)+\frac{I_d}{1-t_{n-1}}\right\|_{
 \mathrm F}\\
 &\qquad\leq\frac{1}{t_{n-1}(1-t_{n-1})}
 \left\{\frac{t_{n-1}^2(1-t_n)^2}
 {256L(t_n-t_{n-1})^2\{d+12\log(N+1)\}}
 \cdot4L\{d+12\log(N+1)\}\right\}^{1/2}\\
 &=\frac{1-t_n}{8(1-t_{n-1})(t_n-t_{n-1})}.
\end{align*}
Consequently, we know from Assumption~\ref{ass:schedule} that
\begin{align*}
 (t_n-t_{n-1})\|\nabla_xv_{t_{n-1}}(x)\|_{\mathrm{op}}
 &\leq\frac{t_n-t_{n-1}}{1-t_{n-1}}
       +\frac{1-t_n}{8(1-t_{n-1})}
 \leq\frac{t_n-t_{n-1}}{t_{n-1}(1-t_n)}+\frac18
 \leq\frac58.
\end{align*}
In particular,
$I_d+\theta(t_n-t_{n-1})\nabla_xv_{t_{n-1}}(x)$ is nonsingular
for every $0\leq\theta\leq1$. Its determinant is continuous in
$\theta$ and equals one at $\theta=0$, so
$\det\nabla_xT_n(x)>0$. This proves \eqref{eq:good-jacobian}.
\end{proof}

\subsection{Proof of Lemma~\ref{lem:source-estimates}}\label{app:source-moments}
\paragraph{Proof of \eqref{eq:source-score-mgf}.}
For every $z,v\in\mathbb R^d$, the fundamental theorem of calculus shows that
\begin{align*}
	&\log \psource(z+v)-\log \psource(z)-v^\top s(z)=\int_0^1 v^\top[s(z+\theta v)-s(z)]\,{\rm d}\theta.
\end{align*}
Using the Lipschitz property of $s$, we obtain
\begin{align}
	\left|\log \psource(z+v)-\log \psource(z)-v^\top s(z)\right|
	&\leq\int_0^1\|v\|_2 L\theta\|v\|_2\,{\rm d}\theta=\frac L2\|v\|_2^2,\label{ineq:remainder_term}
\end{align}
which further implies that
\begin{align*}
	\psource(z)e^{v^\top s(z)} \leq e^{L\|v\|_2^2/2}\psource(z+v).
\end{align*}
Integrating and then changing variables by $z'=z+v$ yields
\begin{align}\label{eq:directional-score-mgf-bound}
	\mathbb E[e^{v^\top s(Z)}]
	\leq e^{L\|v\|_2^2/2}\int \psource(z+v)\,{\rm d}z
	=e^{L\|v\|_2^2/2}\int \psource(z')\,{\rm d}z'
	=e^{L\|v\|_2^2/2}.
\end{align}

Next, we derive the mean and covariance statements.
Fix $v\in\mathbb R^d$. Since
\[
|v^\top s(Z)|\leq e^{v^\top s(Z)}+e^{-v^\top s(Z)},
\]
we know from \eqref{eq:directional-score-mgf-bound} that $\mathbb E[|v^\top s(Z)|]<\infty$.
For every $\lambda>0$, Jensen's inequality and \eqref{eq:directional-score-mgf-bound} together reveal that
\begin{align*}
	\lambda\mathbb E[v^\top s(Z)]
	&\leq\log\mathbb E[e^{\lambda v^\top s(Z)}]
	\leq\frac{L\lambda^2\|v\|_2^2}{2},\\
	-\lambda\mathbb E[v^\top s(Z)]
	&\leq\log\mathbb E[e^{-\lambda v^\top s(Z)}]
	\leq\frac{L\lambda^2\|v\|_2^2}{2}.
\end{align*}
Dividing by $\lambda$ and letting $\lambda\downarrow0$ shows that
$\mathbb E[v^\top s(Z)]=0$. Taking $v$ to be each coordinate vector
proves $$\mathbb E[s(Z)]=0.$$

For every $\lambda>0$, adding \eqref{eq:directional-score-mgf-bound} for
$\lambda v$ and $-\lambda v$ leads to
\begin{align*}
	\mathbb E\left[\frac{2(\cosh(\lambda v^\top s(Z))-1)}{\lambda^2}\right] &=  \frac{1}{\lambda^2}\left(\mathbb E[e^{\lambda v^\top s(Z)}] + \mathbb E[e^{-\lambda v^\top s(Z)}] - 2\right)\leq\frac{2(e^{L\lambda^2\|v\|_2^2/2}-1)}{\lambda^2}.
\end{align*}
Note that $\frac{2(\cosh(\lambda v^\top s(Z))-1)}{\lambda^2}$ is nonnegative and converges pointwise to
$(v^\top s(Z))^2$ as $\lambda\downarrow0$. Therefore, Fatou's lemma
implies
\begin{align*}
	\mathbb E[(v^\top s(Z))^2]
	&\leq\liminf_{\lambda\downarrow0}
	\mathbb E\left[\frac{2(\cosh(\lambda v^\top s(Z))-1)}{\lambda^2}\right] \leq\lim_{\lambda\downarrow0}
	\frac{2(e^{L\lambda^2\|v\|_2^2/2}-1)}{\lambda^2}
	=L\|v\|_2^2.
\end{align*}
Thus every coordinate of $s(Z)$ has a finite second moment, and
$v^\top\mathbb E[s(Z)s(Z)^\top]v\leq L\|v\|_2^2$ for every $v$.
This proves all three statements in \eqref{eq:source-score-mgf}.

\paragraph{Proof of \eqref{eq:source-score-norm-mgf}.}
For every fixed $z$, recognizing that
\[
-\frac{\|g\|_2^2}{2}+\frac{g^\top s(z)}{\sqrt{2L}}
=-\frac12\left\|g-\frac{s(z)}{\sqrt{2L}}\right\|_2^2
+\frac{\|s(z)\|_2^2}{4L},
\]
we obtain
\[
\frac1{(2\pi)^{d/2}}\int_{\mathbb R^d}
e^{-\|g\|_2^2/2+g^\top s(z)/\sqrt{2L}}\,{\rm d}g
=e^{\|s(z)\|_2^2/(4L)}.
\]
Then Tonelli's theorem and
\eqref{eq:source-score-mgf} together show that
\begin{align*}
	\mathbb E\bigl[e^{\|s(Z)\|_2^2/(4L)}\bigr]
	&=\frac1{(2\pi)^{d/2}}\int_{\mathbb R^d}
	e^{-\|g\|_2^2/2}\mathbb E\left[e^{g^\top s(Z)/\sqrt{2L}}\right]\,{\rm d}g\\
	&\leq\frac1{(2\pi)^{d/2}}\int_{\mathbb R^d}
	e^{-\|g\|_2^2/2}e^{\|g\|_2^2/4}\,{\rm d}g\\
	&=\frac1{(2\pi)^{d/2}}\prod_{j=1}^d
	\int_{\mathbb R}e^{-g_j^2/4}\,{\rm d}g_j
	=\frac{(4\pi)^{d/2}}{(2\pi)^{d/2}}=2^{d/2}.
\end{align*}
Therefore, \eqref{eq:source-score-norm-mgf} holds.

\paragraph{Proofs of \eqref{eq:source-Stein} and \eqref{eq:scale-lower-bounds}.} Eqn.~\eqref{eq:source-Stein} is a direct consequence of \cite{rao1945information} and we prove it for completeness. More specifically, we let
\begin{align*}
	\phi(z;\theta)=\psource(z-\theta),
	\qquad
	T(z)=z-\mathbb E[Z].
\end{align*}
Then we have
\begin{align}\label{eq:unbiased_T}
	\int_{\mathbb R^d}T(z)\phi(z;\theta)\,{\rm d}z=\theta
\end{align}
and
\begin{align*}
	\nabla_\theta\log\phi(z;\theta)=-s(z-\theta).
\end{align*}
Moreover, since $\mathbb{E}[\|Z\|_2^2] < \infty$ and
\begin{align*}
	\mathbb{E}[\|s(Z)\|_2^2] &\leq 2\left(\|s(\mathbb{E}[Z])\|_2^2 + \mathbb{E}[\|s(Z) - s(\mathbb{E}[Z])\|_2^2]\right)\\
	&\leq 2\left(\|s(\mathbb{E}[Z])\|_2^2 + L^2\mathbb{E}[\|Z - \mathbb{E}[Z]\|_2^2]\right)\\
	&< \infty,
\end{align*}
we know that for every $\theta \in \mathbb{R}^d$ and $i,j \in\{1, \dots, d\}$, the integral on the right below is absolutely convergent.
For the location family and affine $T$, the fundamental theorem of calculus
and Fubini's theorem justify differentiating the expectation, and
\begin{align*}
	\frac{\partial}{\partial\theta_j}\int_{\mathbb{R}^d}T_i(z)\phi(z;\theta)\,{\rm d}z = \int_{\mathbb{R}^d}T_i(z)\frac{\partial \phi(z;\theta)}{\partial\theta_j}\,{\rm d}z = \delta_{ij}.
\end{align*}
Equivalently, we have
\begin{align*}
	\int_{\mathbb R^d}
	T(z)\bigl(\nabla_\theta\log\phi(z;\theta)\bigr)^\top
	\phi(z;\theta)\,{\rm d}z=I_d,
\end{align*}
Evaluating the identity at $\theta=0$ therefore gives
\begin{align*}
	\mathbb E\left[(Z-\mathbb E[Z])s(Z)^\top\right]=-I_d.
\end{align*}

By virtue of the Cauchy--Schwarz inequality, we have
\begin{align*}
	d&=\left|\mathbb E[(Z-\mathbb E[Z])^\top s(Z)]\right| \leq \mathbb E[\|Z-\mathbb E[Z]\|_2\|s(Z)\|_2]
	\leq\big(\mathbb E\bigl[\|Z-\mathbb E[Z]\|_2^2\bigr]\big)^{1/2}\big(\mathbb E\bigl[\|s(Z)\|_2^2\bigr]\big)^{1/2}.
\end{align*}
In addition, \eqref{eq:source-score-mgf} shows that
\begin{align*}
	\mathbb E\bigl[\|s(Z)\|_2^2\bigr]
	= \operatorname{Tr}(\mathbb E[s(Z)s(Z)^\top]) \leq Ld.
\end{align*}
Putting the previous two inequalities together, we arrive at
\eqref{eq:scale-lower-bounds}.

%



\bibliographystyle{plainnat}
\bibliography{reference}

\end{document}